\PassOptionsToPackage{sort,compress}{natbib}

\documentclass[11pt,a4paper]{gdm_format}
\usepackage[authoryear,sort&compress,round]{natbib}
\let\cite\citep

\usepackage{graphicx}
\usepackage{multirow}
\usepackage{amsmath,amssymb,amsfonts}
\usepackage{amsthm}
\usepackage{mathrsfs}
\usepackage[title]{appendix}
\usepackage{xcolor}
\usepackage{textcomp}
\usepackage{booktabs}
\usepackage{algorithm}
\usepackage{algorithmicx}
\usepackage{algpseudocode}
\usepackage{listings}
\usepackage{caption}
\usepackage{subcaption}
\usepackage{float}
\usepackage{hyperref}

\title{NeuroAgent: LLM Agents for Multimodal Neuroimaging Analysis and Research}

\author[1,2]{Lujia Zhong$^{\dagger}$}
\author[1]{Yihao Xia$^{\dagger}$}
\author[1,2]{Jianwei Zhang$^{\dagger}$}
\author[1,3]{Shuo huang}
\author[1,2]{Jiaxin Yue}
\author[1,2]{Mingyang Xia}
\author[1,2,3]{Yonggang Shi}
\author[ ]{for
the Alzheimer’s Disease Neuroimaging Initiative
}

\affil[1]{Stevens Neuroimaging and Informatics Institute, Keck School of Medicine, University of Southern California}
\affil[2]{Ming Hsieh Department of Electrical and Computer Engineering, Viterbi School of Engineering, University of Southern California}
\affil[3]{Alfred E. Mann Department of Biomedical Engineering, Viterbi School of
Engineering, University of Southern California}
\affil[ ]{$^{\dagger}$These authors contributed equally to this work.}


\begin{document}

\begin{abstract}
Multimodal neuroimaging analysis often involves complex, modality-specific preprocessing workflows that require careful configuration, quality control, and coordination across heterogeneous toolchains. Beyond preprocessing, downstream statistical analysis and disease classification commonly require task-specific code, evaluation protocols, and data-format conventions, creating additional barriers between raw acquisitions and reproducible scientific analysis. We present NeuroAgent, an LLM-driven agentic framework that automates key preprocessing and analysis steps for heterogeneous neuroimaging data, including sMRI, fMRI, dMRI, and PET, and supports interactive downstream analysis through natural-language queries. NeuroAgent employs a hierarchical multi-agent architecture with a feedback-driven Generate-Execute-Validate engine: agents autonomously generate executable preprocessing code, detect and recover from runtime errors, and validate output integrity. We evaluate the system on 1,470 subjects pooled across all ADNI phases (ADNI1, ADNI-GO, ADNI2, ADNI3, and ADNI4; CN=1,000, AD=470), where all subjects have sMRI and tabular data, with subsets also having Tau-PET ($n=469$), fMRI ($n=278$), and DTI ($n=620$). Pipeline ablation studies across multiple LLM backends show that capable models reach up to 100\% intent-parsing accuracy, with the strongest backend (Qwen3.5-27B) reaching 84.8\% end-to-end preprocessing step correctness. Automated recovery limits manual intervention to edge cases where human review is required via the Human-In-The-Loop (HITL) interface. For Alzheimer's Disease (AD) classification using automatically preprocessed multimodal data, our agent ensemble achieves an AUC of 0.9518 with four modalities, outperforming all single-modality baselines. These results show that NeuroAgent can reduce the manual effort required for neuroimaging preprocessing and enables end-to-end automated analysis pipelines for neuroimaging research.
\end{abstract}

\maketitle

\section{Introduction}\label{sec:intro}

Neuroimaging analysis increasingly links mechanistic understanding in neuroscience to evidence used in clinical neurology. As research moves toward precision medicine and large cohorts, the analytic target is shifting from qualitative reads and gross structural abnormalities to quantitative, subject-level multi-modality biomarkers that jointly capture anatomy, tissue microstructure, brain function, and molecular pathology. Multimodal neuroimaging is therefore central: clinically meaningful alterations rarely align with a single contrast or biological scale \cite{loftus2023multimodality}. Structural MRI (sMRI) quantifies macroscopic morphology (for example, cortical thinning and hippocampal atrophy). Diffusion tensor imaging (DTI) probes white matter microstructure via diffusion-derived indices \cite{basser1994dti}. Functional MRI (fMRI) infers large-scale network organization from hemodynamic signals \cite{smith2013resting}. Positron emission tomography (PET) adds molecular readouts such as protein deposition and glucose metabolism. Together, these modalities provide complementary information that single-modality pipelines cannot fully recover.

Alzheimer's disease (AD) makes this multimodal requirement concrete. The 2024 NIA-AA criteria define AD biologically through amyloid, tau, and neurodegeneration biomarkers, explicitly linking molecular pathology to network dysfunction, microstructural injury, and atrophy \cite{jack2024revised,jack2018niaaa}. Across the disease continuum, these processes unfold with stage-dependent sensitivity: PET may detect amyloid or tau burden before pronounced structural loss, whereas sMRI often strengthens as neurodegeneration becomes anatomically evident. Integrating sMRI, fMRI, Tau PET, and even tabular information (demographics, vitals, CSF biomarkers, etc.) thus aligns neuroimaging analysis with early detection, staging, differential diagnosis, and assessment of treatment response \cite{odusami2023machine}.

Despite the diagnostic promise of multimodal cohorts, a major practical bottleneck is often the conversion of raw acquisitions into analysis-ready images through reliable, modality-specific preprocessing. Echo-planar functional data are particularly sensitive to head motion, which can induce structured artifacts in downstream connectivity estimates \cite{power2012motion}; anatomical and diffusion acquisitions are additionally affected by intensity non-uniformity, partial-volume effects, and susceptibility-induced geometric distortions. Meaningful inference therefore depends on long, modality-specific chains that typically include brain extraction or tissue classification, within-subject realignment, bias-field correction, slice-time correction when temporal ordering matters, and nonlinear spatial normalization to a stereotaxic template (for example, MNI152) \cite{smith2002bet,jenkinson2012fsl}. Historically, these chains were assembled from distinct ecosystems, FSL \cite{jenkinson2012fsl}, SPM \cite{ashburner2005spm}, ANTs \cite{avants2011ants}, and FreeSurfer \cite{fischl2012freesurfer}, with different conventions for file formats, header handling, and coordinate frames. Orchestration frameworks such as Nipype and reproducible presets such as fMRIPrep reduce manual wiring \cite{gorgolewski2011nipype,esteban2019fmriprep}, however, deployments on heterogeneous or incomplete datasets may still require additional checks, parameter adjustment, or recovery procedures outside the default workflow. Large studies therefore still lean on additional quality control and parameter tuning \cite{esteban2019mriqc}, which requires more manual work, limits throughput, and introduces rater variance. This operational fragility sits alongside broader evidence that, once preprocessing and modelling choices diverge, analysts can draw markedly different conclusions from shared neuroimaging data \cite{botviniknezer2020variability}.

Parallel to these infrastructure challenges, deep learning has become a dominant tool in medical image analysis \cite{shen2017deep}. Convolutional neural networks and vision transformers can reach strong benchmark accuracy on tasks such as tumor segmentation and disease classification, yet putting them to work on a new dataset, scanner, or scientific question rarely amounts to running a single pre-trained model: domain shifts across sites and acquisition protocols, task shifts from classification to regression or segmentation, silent failures of upstream preprocessing, and limited interpretability of end-to-end predictions all push the burden back onto researchers. In practice this manifests as substantial expert time spent on per-cohort harmonization, per-task code rewrites, manual quality control, and ROI-level post-hoc interpretation. The primary cost of multimodal neuroimaging analysis is therefore not raw compute but human work, and it is this expert labor, rather than any single algorithmic limitation, that an autonomous analysis system must absorb.

Recent advances in Large Language Models (LLMs) offer potential to absorb much of this manual burden. Early applications in medicine focused on textual reasoning, such as summarizing electronic health records or answering medical-licensing-style questions, where frontier models began to approach expert-level performance \cite{singhal2023medpalm}. The latest generation has demonstrated agentic capabilities \cite{wang2024survey}. Unlike passive text generators, AI agents perceive a digital environment, maintain stateful memory, decompose abstract goals into sub-tasks, and execute external tools \cite{yao2022react,schick2023toolformer}. This mimics the reasoning process of a human data analyst: inspecting the data, selecting the appropriate tool for the specific modality, verifying the output, and correcting errors if the result is suboptimal. For neuroimaging, this matters at multiple stages: agents can configure and adapt preprocessing pipelines to heterogeneous inputs, perform in-pipeline quality control by reading logs and intermediate outputs and re-issuing tool calls when checks fail, and, once data are preprocessed, drive post-hoc analyses such as group-wise statistics, ROI-level regression against clinical variables, and natural-language reporting. The transition from rigid script-based automation to reasoning-based autonomy thus lets the system act on the \textit{context} of the data it is processing across the full lifecycle, rather than only on fixed instructions for a single stage.

In this work, we present NeuroAgent, an LLM agent framework that automates the full lifecycle of multimodal neuroimaging processing and analysis. NeuroAgent functions as an intelligent orchestrator rather than a static execution engine. At its core is a Planning Module powered by an LLM, which interacts with a curated library of domain-specific tools wrapping established algorithms (e.g.,\ ANTs for registration and FSL for extraction). Given a high-level natural-language prompt and directories of raw medical images, NeuroAgent autonomously scans the headers to identify modalities (sMRI, DTI, fMRI, and PET). It then constructs a dependency graph for preprocessing and executes the necessary cleaning and normalization steps. The system also incorporates a self-correction mechanism. If a tool fails (e.g.,\ a registration divergence), the agent parses the error log, adjusts parameters (e.g.,\ increasing the search radius or changing the cost function), and retries the operation without human intervention.
We evaluate NeuroAgent on data pooled across all longitudinal ADNI phases (ADNI1, ADNI-GO, ADNI2, ADNI3, and ADNI4; $n=1{,}470$; CN $n=1{,}000$, AD $n=470$); every subject has sMRI and tabular covariates, with nested subsets additionally providing Tau-PET ($n=469$), resting-state fMRI ($n=278$), and DTI ($n=620$).
We organize the evaluation in three stages, mirroring how the system would be used in practice. First, \emph{pipeline testing}: across LLM backends, ablations on intent parsing, preprocessing code generation, and data integration report up to $100\%$ intent-parsing accuracy and $84.8\%$ end-to-end preprocessing step correctness for the strongest backend (Qwen3.5-27B), with automated recovery so that manual handling is reserved for edge cases escalated through the HITL interface. Second, \emph{post-interactive group analysis}: on automatically preprocessed sMRI and DTI cohorts, the agent reproduces clinically expected effects, including AD-associated cortical thinning and age-related decline of fractional anisotropy, indicating that pipeline outputs are consistent with established clinical insights. Third, \emph{downstream classification}: on automatically preprocessed four-modality inputs, an agent ensemble reaches ROC-AUC $0.9518$ for AD vs.\ CN classification and exceeds all single-modality baselines.

\paragraph*{Contributions.} (1) We introduce a unified autonomous preprocessing engine capable of handling heterogeneous neuroimaging inputs (sMRI, DTI, fMRI, PET). By abstracting the complexity of underlying tools (FreeSurfer, FSL, MRtrix3, Elastix), the system autonomously resolves cross-modality dependencies and executes complete preprocessing pipelines without manual parameter tuning. (2) We present a ``Generate-Execute-Validate'' mechanism that supports pipeline execution through automated error recovery, and ablate it across multiple LLM backends to quantify intent-parsing accuracy, preprocessing step correctness, and data-integration fidelity, establishing a systematic benchmark for LLM-driven neuroimaging pipeline agents. (3) We show that NeuroAgent connects automated preprocessing with post-interactive group analysis and downstream classification on the ADNI dataset: the system autonomously preprocesses multimodal data, performs natural-language-driven group-level statistics, and supports AD classification with an agent ensemble achieving AUC $0.9518$, all from a Human-in-the-Loop interface that lets researchers supervise, intervene on escalated failures, and visualize results interactively.

\section{Related Work}\label{sec:related}
\subsection{Neuroimaging Pipelines and Automation}
Neuroimaging analysis has transitioned from manual, variable scripting toward standardized and automated orchestration. Historically, the field faced high analytical variability, as demonstrated by the Neuroimaging Analysis Replication and Prediction Study (NARPS) \cite{botviniknezer2020variability}. In that study, seventy independent teams analyzed the same fMRI dataset and no two teams chose identical workflows; variation across preprocessing software, smoothing kernels, model specification, and statistical thresholds produced measurably divergent hypothesis test results. To address these challenges, the community developed workflow systems such as Nipype, which provides a uniform interface to legacy software including AFNI, ANTs, FreeSurfer, FSL, and SPM \cite{gorgolewski2011nipype}.

The NiPreps ecosystem, particularly fMRIPrep, represents a major advance in standardized preprocessing and reproducible execution \cite{esteban2019fmriprep}. By using BIDS metadata, these pipelines substantially reduce manual parameterization and improve interoperability across datasets. However, these systems are primarily designed around predefined workflows: they work best when inputs follow expected structures, metadata are available, and preprocessing logic can be specified in advance. In practice, heterogeneous clinical datasets, non-standard directory structures, missing metadata, or tool failures often require manual intervention outside the scope of static pipelines. NeuroAgent is motivated by this remaining gap between reproducible workflow execution and adaptive, context-aware pipeline construction.

\subsection{Large Language Models in Medical Imaging}
Large Language Models (LLMs) have shifted medical informatics toward clinical reasoning, report generation, and multimodal interpretation. Specialized models such as Med-PaLM show strong performance on medical question answering and decision-support benchmarks \cite{singhal2023medpalm}, while open medical foundation models such as MedGemma improve accessibility for multimodal medical research \cite{sellergren2025medgemma}. These advances suggest that LLMs are increasingly useful as high-level semantic interfaces for medical data.

However, the strengths of medical LLMs do not directly translate into reliable scientific workflow execution. Neuroimaging pipelines require deterministic tool invocation, strict file handling, parameter consistency, and recovery from runtime or data-quality failures. In these settings, LLMs still struggle with code-heavy, tool-dependent workflows and can exhibit non-robust reasoning when invoked as standalone executors. A recent benchmark on brain MRI interpretation reported that GPT-4o declined to respond to roughly 28\% of slices after three attempts, including some normal scans, while Grok and Gemini answered every slice on the first attempt \cite{sozer2025mri}. This indicates that current general-purpose LLMs can be unreliable when called directly on imaging inputs. Computational scale and privacy also remain practical barriers. Compression methods such as SparseLLM reduce deployment cost, but efficient deployment alone does not close the reliability gap between language understanding and trustworthy execution in neuroimaging pipelines \cite{bai2024sparsellm}. For this reason, LLMs are better viewed as planners, controllers, or interfaces layered over deterministic tools rather than as standalone engines for medical image analysis.

\subsection{Autonomous Agents for Medical Imaging}
Autonomous agents operate within a continuous loop of perception, reflection, and action, using external tools to solve multi-stage problems. Frameworks such as AutoGen provide the scaffolding for multi-agent collaboration, while ReAct formalizes the coupling between reasoning traces and tool-executing actions \cite{wu2023autogen,yao2022react}. These ideas are particularly relevant to scientific workflows, where the ability to plan, execute, inspect intermediate outputs, and revise actions is more valuable than single-pass generation.

In medical imaging, recent systems demonstrate several complementary directions. AgentMRI applies agentic control to MRI reconstruction \cite{sajua2025agentmri}. AURA and TissueLab move toward more interactive medical imaging agents by combining tool use, explanation, and iterative analysis \cite{fathi2025aura,li2025tissuelab}. MACRO further studies self-evolving medical imaging agents that acquire reusable composite skills from prior trajectories \cite{fan2026macro}. At the evaluation level, ReX-MLE and MedMASLab emphasize that agentic medical systems also require dedicated end-to-end benchmarks and orchestration frameworks rather than isolated task metrics \cite{kenia2025rexmle,qian2026medmaslab}. Domain-specialized neuroscience LLMs such as BrainGPT show that expert priors can improve performance on neuroscience reasoning tasks \cite{luo2024brainbench}.

For Alzheimer-focused applications, ADAgent, AD-CARE, and AD-Reasoning extend agentic reasoning toward multimodal diagnosis, guideline grounding, and clinically structured outputs \cite{hou2025adagent,hou2026adcare,chen2026adreasoning}. These systems make important progress on individual stages such as reconstruction, reporting, benchmarking, or diagnosis, but they typically operate on data that is already prepared and do not span the full multimodal preprocessing and downstream analysis lifecycle. NeuroAgent is positioned to fill this gap by covering the lifecycle from raw multimodal neuroimaging preprocessing to downstream analysis, with explicit support for dependency resolution, reflective error recovery, and human-in-the-loop escalation.

\section{Methodology}\label{sec:methods}

\subsection{Framework Overview}
\begin{figure}[t]
    \centering
    \includegraphics[width=\linewidth]{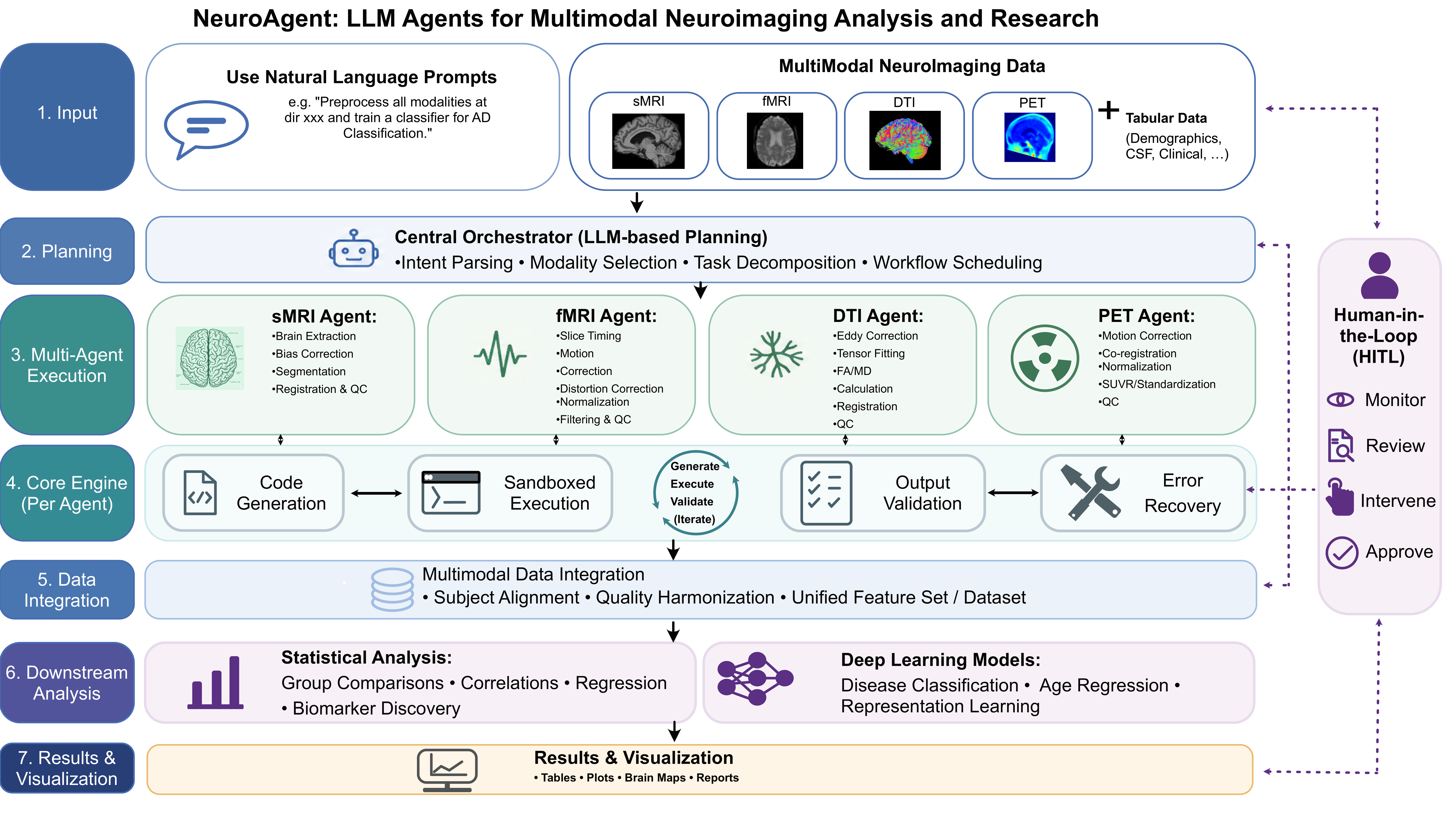}
    \caption{\textbf{NeuroAgent Framework Overview.} The system comprises a Central Orchestrator (planning), Specialized Modality Agents (execution), and a Feedback-Driven ``Generate-Execute-Validate'' engine that enables reflective self-correction. A Human-in-the-Loop interface allows researchers to supervise and intervene at critical decision points.}
    \label{fig:framework}
\end{figure}

We propose NeuroAgent, a hierarchical, autonomous multi-agent framework designed to orchestrate the complete lifecycle of neuroimaging analysis, from raw data preprocessing to downstream scientific tasks. As illustrated in Figure~\ref{fig:framework}, the system mimics the organizational structure of a research pipeline, separating high-level cognitive planning from low-level technical execution. The architecture is composed of three primary components: a Central Orchestrator, a set of Specialized Modality Agents, and a Feedback-Driven Execution Engine.

\subsubsection{Hierarchical Multi-Agent Architecture}
To handle the complexity of heterogeneous medical data, the framework employs a two-tier agent hierarchy. At the top level, the \textbf{Central Orchestrator} is the Planning Module: it parses natural language research goals (e.g., "Analyze functional connectivity changes in AD patients"), decomposes abstract intents into a structured dependency graph, and dispatches sub-tasks to the appropriate domain experts. At the execution level, the framework instantiates \textbf{Specialized Modality Agents} for each data modality (e.g., sMRI, fMRI, PET, dMRI). These agents encapsulate domain-specific knowledge bases and toolchains, so that processing decisions (such as registration templates or denoising parameters) are appropriate for the specific imaging type.

\subsubsection{The Feedback-Driven Execution Engine}
A core design decision in NeuroAgent is its departure from static, linear scripts. Instead, it uses an iterative "Generate-Execute-Validate" engine that bridges the agents' reasoning and the operating system, drawing on recent reasoning-and-tool-use approaches for LLM agents \cite{yao2022react,schick2023toolformer}. This engine operates on a closed-loop mechanism:
\begin{enumerate}
    \item \textbf{Contextual Code Generation:} Agents generate executable Python scripts to call external neuroimaging libraries based on the current data state.
    \item \textbf{Sandboxed Execution \& Self-Correction:} The engine executes these scripts in a controlled environment. If an execution error occurs (e.g., a tool failure), the runtime error logs are captured and fed back to the agent, triggering an automatic debugging and regeneration cycle.
    \item \textbf{Structural Integrity Validation:} Beyond runtime errors, the engine enforces strict quality assurance by verifying the integrity of the output file structure against predefined schemas (e.g., BIDS compliance). A step is marked complete only when the output directory satisfies these structural constraints, reducing the risk that incomplete or malformed outputs propagate to downstream stages.
\end{enumerate}

\subsubsection{Workflow Orchestration}
The system orchestrates the research pipeline through four sequential but adaptive phases:
\begin{enumerate}
    \item \textbf{Task Distribution:} The Orchestrator analyzes the input data and research goal to determine the necessary modalities and analysis tasks, constructing a directed acyclic graph (DAG) of dependencies (e.g., prioritizing structural MRI processing to support fMRI registration).
    \item \textbf{Parallel Preprocessing:} Specialized agents concurrently process their respective data streams. This phase utilizes the execution engine's self-correction capabilities to adaptively handle data heterogeneity.
    \item \textbf{Data Integration:} Upon the completion of preprocessing, the system aggregates disparate outputs into a unified, multi-modal dataset. This phase automatically aligns subjects across modalities and generates a consolidated manifest for downstream analysis.
    \item \textbf{Task Processing:} In the final phase, the system configures and executes the specific analytical models (e.g., classification, regression, or group analysis) utilizing the organized data.
\end{enumerate}

To support reliability in clinical contexts, the framework also incorporates a "Human-in-the-Loop" mechanism, allowing researchers to intervene or approve critical decisions (such as task allocation or validation failures) before the system proceeds.

\subsection{The Planning Module}
The Planning Module serves as the core of the NeuroAgent framework, responsible for translating high-level natural language directives into a structured, executable workflow. Powered by a Large Language Model (LLM), this module functions as a semantic parser and a strategic scheduler, ensuring that the downstream computational steps align with the user's scientific intent.

\subsubsection{Intent Parsing}
Upon receiving a user prompt (e.g., "Train a 3D CNN to classify Alzheimer's Disease using Tau-PET images"), the Planning Module first performs intent analysis to extract two critical sets of entities: the required \textit{Data Modalities} and the \textit{Target Tasks}. Instead of relying on rigid keyword matching, the module utilizes the semantic reasoning capabilities of the LLM to map the user's request to the available specialized agents. For instance, if a user requests an analysis involving "functional connectivity," the module infers the necessity of the fMRI agent. Similarly, it identifies the specific downstream analysis required, distinguishing between tasks such as diagnosis (classification), continuous variable prediction (regression), or statistical correlation analysis.

\subsubsection{Workflow Construction}
A core function of the Planning Module is the automated resolution of cross-modality dependencies. Neuroimaging workflows often have prerequisites that are not explicitly stated by the user. The Planning Module enforces these logical constraints through a dynamic dependency graph. For example, processing functional (fMRI) or diffusion (dMRI) data typically requires a high-resolution structural reference (T1-weighted MRI) for anatomical registration and segmentation. Even if the user's prompt only mentions "fMRI analysis," the Planning Module detects this dependency and injects the structural MRI preprocessing steps into the workflow before the functional pipeline. The specialized agents then receive correctly co-registered inputs without requiring manual specification from the user.

\subsubsection{Task Identification}
Beyond data preparation, the Planning Module addresses the user's analytical goals, such as model training or statistical inference. By interpreting the research objective given in the user prompt, the module schedules the appropriate \textit{Task Processing} phase. The Planning Module selects a solver for the identified task (e.g., a 3D-CNN solver for disease diagnosis or a correlation solver for biomarker analysis) and defines the input requirements for the Data Integration phase. The output of the preprocessing agents is then not just cleaned data but data formatted and harmonized to feed the input layer of the subsequent machine learning models or statistical tools. This end-to-end planning capability lets the framework move from raw DICOM conversion to the final training metrics or diagnostic reports.

\subsection{Tool Library and Integration}
\subsubsection{Preprocessing Tools}
The framework encapsulates a suite of domain-specific neuroimaging tools, standardized into modular Python interfaces. Rather than reinventing established algorithms, NeuroAgent wraps widely used software packages (including FreeSurfer~\cite{fischl2012freesurfer}, FSL~\cite{jenkinson2012fsl}, and Elastix~\cite{klein2010elastix}), so that preprocessing pipelines follow validated scientific protocols. The tool library is organized by imaging modality.

\noindent\textbf{Data Ingestion and Standardization.} To handle raw clinical data, the system includes a unified ingestion module capable of converting heterogeneous DICOM series into NIfTI format. This module wraps the \texttt{dcm2niix} utility, automatically handling BIDS sidecar generation and gzip compression to ensure standardized metadata availability for downstream agents.

\noindent\textbf{Structural MRI (sMRI) Pipeline}
The structural backbone of the framework is a wrapper for FreeSurfer's \texttt{recon-all} stream. This module automates the full cortical reconstruction process, including motion correction, skull stripping, and subcortical segmentation. Beyond standard morphological metrics, the agent runs additional segmentation routines (specifically \texttt{gtmseg} for anatomical region definition and \texttt{segmentBS} for brainstem sub-segmentation) to support fine-grained region-of-interest (ROI) analysis.

\noindent\textbf{Functional MRI (fMRI) Pipeline}
The fMRI processing chain uses a hybrid approach, integrating FreeSurfer and FSL to address both spatiotemporal artifacts and physiological noise. For \emph{spatiotemporal correction}, the module uses \texttt{preproc-sess} to perform slice-timing correction, motion correction, and surface projection. It generates data in both standard (fsaverage) and native spaces, applying 5mm FWHM smoothing to enhance signal-to-noise ratio. For \emph{denoising and connectivity}, the system incorporates an FSL-based denoising routine. It performs tissue segmentation (via FAST) to extract White Matter (WM) and Cerebrospinal Fluid (CSF) signals, which are then regressed out as nuisance variables. The cleaned time series are then transformed into connectivity matrices for functional network analysis.

\noindent\textbf{Diffusion MRI (dMRI) Pipeline}
The diffusion MRI (dMRI) processing chain uses a custom, automated Python-based pipeline that integrates tools from FSL, DIPY, and MRtrix3. This workflow handles heterogeneous acquisition protocols autonomously while interfacing with FreeSurfer outputs for anatomically informed analysis.

The agent begins by extracting acquisition parameters (such as phase-encoding direction and total readout time) directly from BIDS metadata to select a distortion correction strategy. For datasets containing reverse phase-encoded $b=0$ images, the pipeline executes FSL \texttt{topup} to estimate and correct susceptibility-induced distortions. All diffusion volumes then undergo motion and eddy current correction via FSL \texttt{eddy}, which incorporates susceptibility field estimates and enables outlier slice replacement and Jacobian modulation to mitigate signal dropout artifacts. Brain masking is performed using FSL \texttt{bet}, and after correction, the diffusion gradient vectors (b-vecs) are rotated to maintain alignment with the corrected image data.

For microstructural modeling, the system adapts to the available data shells. Diffusion Tensor Imaging (DTI) metrics (fractional anisotropy (FA), mean diffusivity (MD), axial diffusivity (AD), and radial diffusivity (RD)) are derived using single-shell subsets (typically $b=1000$ s/mm$^2$). To resolve complex fiber configurations, the pipeline estimates Fiber Orientation Distributions (FODs) using constrained spherical deconvolution (CSD) in MRtrix3. The agent selects between single-shell and multi-shell multi-tissue CSD based on the input shells, modeling spherical harmonics up to $l_{\max}=8$.

Integration with structural data is achieved by registering FreeSurfer-derived anatomical parcellations to diffusion space using boundary-based registration (BBR). This enables anatomically constrained tractography (ACT) and region-of-interest (ROI) analysis. The framework then performs whole-brain tractography using dynamic seeding and SIFT2 filtering to normalize streamline weights. These streamlines are mapped to the anatomical parcellations using \texttt{tck2connectome} to generate subject-level structural connectivity matrices.

\noindent\textbf{Positron Emission Tomography (PET) Pipeline}
For molecular imaging, particularly Tau-PET, the framework implements a dynamic frame realignment strategy to correct for subject motion during long acquisition times. The module uses \texttt{Elastix} to perform rigid-body registration, iteratively aligning subsequent frames to the initial reference frame. After coregistration, the module computes a temporally averaged static image, improving signal quality for downstream tracer uptake quantification.

\subsubsection{Analysis Tools}
After preprocessing, the framework employs a Model Zoo of analytical tools to extract scientific insights from the preprocessed data. The Planning Module selects an appropriate solver based on the input data structure (e.g., 4D time-series, 3D volumetric maps, or adjacency matrices) and the user's research objective.

\noindent\textbf{Deep Diagnostic Models.} For disease classification, the framework integrates deep learning architectures tailored to specific modalities. For \emph{volumetric classification} (PET/sMRI), the system uses 3D Convolutional Neural Networks (3D-CNNs). The volumetric analysis module ingests preprocessed NIfTI volumes, treating them as voxel grids to learn hierarchical spatial features indicative of pathology distribution (e.g., amyloid or tau deposition patterns). It handles tensor reshaping and batch normalization to prepare the data for high-dimensional inference. For \emph{connectivity-based classification} (fMRI/DTI), the system moves from voxel space to network space. The connectivity analysis module processes correlation matrices derived from the preprocessing phase. It flattens these matrices into feature vectors and uses Multi-Layer Perceptrons (MLPs) or Graph Neural Networks (GNNs) to identify abnormal connectivity fingerprints associated with cognitive decline.

\noindent\textbf{Statistical Analysis and Biomarker Extraction}
Beyond black-box prediction, the framework supports interpretable statistical analysis. It includes utilities to extract Region-of-Interest (ROI) based metrics, such as Standardized Uptake Value Ratios (SUVR) for PET or Fractional Anisotropy (FA) values for DTI, and to perform regression against clinical variables (e.g., MMSE scores, Age, APOE status). This allows researchers to compare deep learning findings against established clinical biomarkers.

\noindent\textbf{Visualization and Automated Reporting}
To support human interpretation, the analysis tools include a visualization engine that operates in a markedly more interactive regime than the largely autonomous preprocessing pipelines. The engine responds to follow-up natural-language requests so that researchers can iteratively probe the data and models for their own analytical needs. For deep learning models, the engine integrates SHAP-based attribution~\cite{lundberg2017unified} to produce voxel- or feature-level importance maps over PET/sMRI volumes and connectivity matrices, helping users localize the regions and edges that drive each prediction. At the cohort level, the engine generates summary reports that aggregate per-subject metrics into group-wise statistics (e.g., AD-vs-CN ROI distributions, FA decline trajectories), produces overlay plots on standard atlases, and exports CSV/HTML artifacts that are linked back to the workflow registry for traceability. Through the HITL interface, users can refine these views on demand, such as requesting alternative parcellations, subgroup splits, or biomarker overlays, turning the analysis phase into an interactive dialogue between the researcher and the agent.

\subsection{Memory and Error Recovery}
To operate reliably in a complex, multi-step research environment, NeuroAgent implements a state management system coupled with a multi-layered error recovery mechanism. To address the safety requirements of clinical research, the framework also integrates a Human-in-the-Loop (HITL) interface, allowing researchers to supervise and guide autonomous agents through a interactive user interface.

\subsubsection{Contextual Memory and State Persistence}
Unlike stateless interaction models, the framework maintains a persistent \textbf{Global Workflow Registry}. This registry tracks the granular status of every phase (Task Distribution, Preprocessing, Integration, Analysis), recording execution times, computational costs, and token usage for each sub-step. This stateful design serves two purposes. First, it enables \emph{cross-modality context sharing}: the registry acts as a blackboard where downstream agents retrieve context from upstream predecessors without direct coupling. For instance, the dMRI Agent can query the registry to locate the specific FreeSurfer output directory generated by the sMRI Agent, ensuring precise anatomical mapping. Second, it provides \emph{resumability}: if a long-running pipeline is interrupted, the system checks the completion status flags in the registry, skips already finalized steps, and resumes execution at the point of interruption.

\subsubsection{Reflective Error Recovery}
The Execution Engine implements a \textbf{Reflective Debugging} mechanism to handle the stochastic nature of code generation. This operates on two levels:

\noindent\textbf{Exception Handling.} When a generated script fails during execution (e.g., due to a syntax error or a library mismatch), the system captures the full stack trace and standard error output. Instead of terminating, the engine feeds this error log back to the agent as a new prompt context. The agent then analyzes the error, modifies the code (e.g., fixing an import path or adjusting a parameter), and re-submits it for execution. This cycle repeats until success or a maximum retry threshold is reached.

\noindent\textbf{Output Validation}
Success in execution does not imply correctness in science. To address this, the system employs a \textbf{Validator} layer. After a script executes successfully, the validator inspects the output directory against expected schemas (e.g., checking for specific BIDS filenames or NIfTI header properties). If the output structure is invalid (e.g., a missing header file), the system rejects the result, constructs a specific feedback message explaining the structural deficiency, and forces the agent to regenerate the processing logic.

\section{Experiments and Results}\label{sec:experiments}

\subsection{Experimental Setup}
We evaluate NeuroAgent on data from the Alzheimer's Disease Neuroimaging Initiative (ADNI), with three complementary evaluations: (1) \textit{pipeline ablation}, measuring how well different LLM backends perform at intent parsing, preprocessing code generation, and data integration; (2) \textit{post-interactive group analysis}, demonstrating that the agent reproduces clinically expected effects on automatically preprocessed sMRI and DTI cohorts via natural-language queries; and (3) \textit{downstream classification}, measuring AD diagnosis accuracy on automatically preprocessed multimodal data.

For preprocessing, the agent autonomously executed the full pipeline for each subject with self-correction for common runtime errors: skull stripping and cortical parcellation via FreeSurfer \texttt{recon-all} for sMRI; eddy current and susceptibility distortion correction via FSL \texttt{topup}/\texttt{eddy}, followed by DTI metric extraction (FA, MD) for dMRI; frame realignment and SUVR computation via Elastix for Tau-PET. Failures not resolved by self-correction were escalated to the HITL interface. Preprocessed data were then assembled by the data integration agent into a unified subject-level table.

For group analysis, the agent operated on automatically preprocessed FreeSurfer cortical-thickness summaries and DTI biomarkers, autonomously matching subjects to modality-specific imaging visits, generating statistical code, executing covariate-adjusted regressions and group comparisons, and summarizing the findings in natural language. The cohorts here are modality-specific and broader than the classification cohort (1{,}601 CN/MCI/AD subjects for cortical thickness; 1{,}137 subjects for DTI); see Section~\ref{sec:post_interactive} for the full setup and results.

For classification, we trained per-modality predictors and combined them through two kinds of ensembles. For volumetric imaging (sMRI, Tau-PET) we trained three 3D CNN backbones implemented in MONAI \cite{cardoso2022monai}: ResNet-18 (33.2M parameters), DenseNet-121 (11.2M), and EfficientNet-B0 (4.7M). Each backbone produces one set of out-of-fold logits per modality. For tabular clinical data and flattened fMRI lower-triangular correlation matrices we used TabPFN, yielding a single set of out-of-fold logits per non-imaging modality. Per-architecture multimodal baseline rows (e.g., ResNet-18 MRI + ResNet-18 Tau-PET) combine the constituent modality logits by simple averaging in logit space. The Agent Ensemble is a stratified 5-fold MLP stacker (two hidden layers of 16 and 8 units, ReLU activations) trained on the union of all raw baseline logits available for the configuration: 3 features for the unimodal sMRI and Tau-PET tables, 6 for sMRI+Tau-PET, and 8 for the four-modality table (3 sMRI + 3 Tau-PET + 1 tabular + 1 fMRI). Note that the Agent Ensemble stacks raw baseline logits directly rather than the per-architecture averaged outputs. For multimodal configurations the evaluation is performed on the union of subjects across modalities, and missing-modality logits are imputed with a neutral value of 0 (probability 0.5) prior to averaging or stacking. Splitting was performed strictly at the \textit{subject level} using 5-fold cross-validation, so that no subject's data appears across train and test folds. All reported classification metrics are averaged across the 5 folds. Class imbalance was addressed via oversampling on training folds only.

\subsection{Dataset for Downstream Classification}
Data were obtained from the ADNI database (\url{adni.loni.usc.edu}). We pooled participants across all longitudinal ADNI phases (ADNI1, ADNI-GO, ADNI2, ADNI3, and ADNI4) \cite{arani2024design,zavaliangos2019adni3,arani2025adni4} for the downstream binary AD vs.\ CN classification task. The final classification cohort comprised 1,470 unique subjects: Cognitively Normal (CN, $n=1{,}000$) and Alzheimer's Disease (AD, $n=470$). All subjects have sMRI and tabular (clinical) data; subsets additionally have Tau-PET ($n=469$), fMRI ($n=278$), or DTI ($n=620$), each nested within the sMRI/tabular cohort. Table~\ref{tab:modality_availability} summarizes modality-specific subject counts for the classification experiments. Demographic characteristics of the classification cohort are provided in Table~\ref{tab:demographics}. Table~\ref{tab:modality_demo_summary} further reports the broader per-modality cohort (including MCI subjects) used in the post-interactive group analyses (Section~\ref{sec:post_interactive}).

\begin{table}[htbp]
\centering
\caption{Demographic characteristics of the study cohort (CN vs.\ AD).}
\label{tab:demographics}
\begin{tabular}{lcc}
\toprule
\textbf{Characteristic} & \textbf{CN} & \textbf{AD} \\
\midrule
No.\ of subjects & 1,000 & 470 \\
Age (years) & $71.9 \pm 7.7$ & $74.8 \pm 7.9$ \\
Sex (Female, \%) & 615 (61.5\%) & 208 (44.3\%) \\
\bottomrule
\multicolumn{3}{l}{\footnotesize CN: Cognitively Normal; AD: Alzheimer's Disease.}\\
\multicolumn{3}{l}{\footnotesize Age: mean $\pm$ standard deviation.}\\
\end{tabular}
\end{table}

\begin{table}[htbp]
\centering
\caption{Subject availability per imaging modality within the AD vs.\ CN classification cohort. All subjects with Tau-PET, fMRI, or DTI data also have sMRI and tabular data. DTI is reported here for completeness; the AD classification experiments in Section~\ref{sec:classification_results} use sMRI, Tau-PET, fMRI, and tabular features.}
\label{tab:modality_availability}
\begin{tabular}{lccc}
\toprule
\textbf{Modality} & \textbf{Total} & \textbf{CN} & \textbf{AD} \\
\midrule
sMRI + Tabular (Clinical) & 1,470 & 1,000 & 470 \\
\quad + Tau-PET & 469 & 406 & 63 \\
\quad + fMRI & 278 & 248 & 30 \\
\quad + DTI & 620 & 563 & 57 \\
\quad + Tau-PET + fMRI + DTI (all five) & 242 & 222 & 20 \\
\bottomrule
\end{tabular}
\end{table}

\begin{table*}[t]
\centering
\caption{Demographic summary of per-modality ADNI cohorts. Each subject is counted once at their first available scan per modality. This broader cohort (including MCI subjects) is the data pool from which the AD vs.\ CN classification cohort (Table~\ref{tab:modality_availability}) and the post-interactive matched cohorts (Section~\ref{sec:post_interactive}) are drawn.}
\label{tab:modality_demo_summary}
\resizebox{\textwidth}{!}{%
\begin{tabular}{lcccccccc}
\toprule
Modality & N & Age, years & Female & Male & CN & MCI & AD & Missing dx \\
\midrule
sMRI & 2{,}240 & 72.9 $\pm$ 7.9 & 1{,}173 (52.4\%) & 1{,}046 (46.7\%) & 1{,}000 (44.6\%) & 770 (34.4\%) & 470 (21.0\%) & 0 (0.0\%) \\
fMRI & 500 & 74.0 $\pm$ 7.5 & 263 (52.6\%) & 237 (47.4\%) & 312 (62.4\%) & 151 (30.2\%) & 37 (7.4\%) & 0 (0.0\%) \\
Tau-PET & 885 & 74.3 $\pm$ 8.1 & 460 (52.0\%) & 425 (48.0\%) & 502 (56.7\%) & 306 (34.6\%) & 77 (8.7\%) & 0 (0.0\%) \\
DTI & 1{,}137 & 72.9 $\pm$ 8.1 & 647 (56.9\%) & 484 (42.6\%) & 707 (62.2\%) & 311 (27.4\%) & 109 (9.6\%) & 10 (0.9\%) \\
\bottomrule
\end{tabular}%
}
\end{table*}

\subsection{Pipeline Ablation Studies}

We systematically evaluated five LLM backends across three pipeline stages: (1) \textit{intent parsing}, correctly identifying the required data modalities and downstream tasks from a natural language prompt; (2) \textit{preprocessing code generation}, producing syntactically correct, tool-valid, and structurally constrained scripts; and (3) \textit{data integration}, assembling subject-level manifests with correct row structure, valid paths, and no duplicates.

\paragraph*{Metric definitions.}
On the \textit{intent parsing} benchmark (18 prompts spanning single- and multi-modality AD diagnosis, age prediction, preclinical detection, and longitudinal progression), we report: \textbf{Modality EM}, the fraction of prompts where the predicted set of required modalities exactly matches the gold label (exact set match); \textbf{Task EM}, the fraction where the predicted downstream task set exactly matches the gold; \textbf{Joint EM}, the fraction satisfying both simultaneously; and \textbf{Invalid Rate}, the fraction where the model failed to produce parseable structured output after three retries.

For \textit{preprocessing code generation} (33 cases across sMRI, Tau-PET, fMRI, and dMRI), we report: \textbf{Syntax}, the generated Python code parses without AST errors; \textbf{Tool}, the expected preprocessing tool is both imported and called; \textbf{In Path}, the referenced input path is grounded to the provided directory tree; \textbf{Out Path}, the referenced output path falls within the expected output directory; \textbf{Step Const.}, step-specific constraints are satisfied (e.g., BIDS-compliant filenames, AP/PA phase-encoding labels, required bookkeeping files such as \texttt{dicom\_dirs.txt}); and \textbf{All Pass}, all five checks pass simultaneously.

In the \textit{data integration} benchmark (8 cases covering single- and multi-modality settings with partial missing data), generated code is executed against a simulated preprocessed-output directory tree and the produced \texttt{final\_data\_list.csv} is compared against a gold reference. We report: \textbf{Row EM}, the exact row-level match after column canonicalization and sorting by (\textit{SubjectID}, \textit{Date}); \textbf{Subject-Date F1}, the $F_1$ score on (\textit{SubjectID}, \textit{Date}) pair coverage (balancing precision and recall over covered subjects); \textbf{Path Validity}, the fraction of non-empty file paths that exist on disk; \textbf{Col. Completeness}, the fraction of required (subject, date, column) triples where the predicted path exactly matches the gold; \textbf{Duplicate-Free}, no duplicate (\textit{SubjectID}, \textit{Date}) rows; and \textbf{All Pass}, all preceding checks satisfied simultaneously.

\begin{table}[t]
\centering
\small
\begin{tabular}{lccccc}
\toprule
Model & \# Prompts & Modality EM & Task EM & Joint EM & Invalid Rate \\
\midrule
ollama-glm-4.7-flash & 18 & 94.4\% & 100.0\% & 94.4\% & 0.0\% \\
ollama-gpt-oss & 18 & 94.4\% & 94.4\% & 88.9\% & 0.0\% \\
ollama-qwen3.5:27b & 18 & 100.0\% & 100.0\% & 100.0\% & 0.0\% \\
ollama-qwen3.5:9b & 18 & 100.0\% & 100.0\% & 100.0\% & 0.0\% \\
ollama-qwen3.5:4b & 18 & 88.9\% & 100.0\% & 88.9\% & 0.0\% \\
ollama-qwen3.5:2b & 18 & 77.8\% & 77.8\% & 77.8\% & 22.2\% \\
ollama-qwen3.5:0.8b & 18 & 0.0\% & 0.0\% & 0.0\% & 100.0\% \\
ollama-qwen3:30b & 18 & 100.0\% & 94.4\% & 94.4\% & 0.0\% \\
ollama-qwen3:14b & 18 & 0.0\% & 0.0\% & 0.0\% & 100.0\% \\
ollama-qwen3:8b & 18 & 0.0\% & 0.0\% & 0.0\% & 100.0\% \\
ollama-qwen3:4b & 18 & 100.0\% & 100.0\% & 100.0\% & 0.0\% \\
ollama-qwen3:1.7b & 18 & 27.8\% & 22.2\% & 22.2\% & 61.1\% \\
\bottomrule
\end{tabular}
\caption{Intent parsing accuracy across prompt benchmarks and model backends for NeuroAgent task distribution. Models are ordered by family (GLM, GPT-OSS, Qwen3.5, Qwen3) with each Qwen family sorted from larger to smaller parameter scale. Exact match (EM) is computed over the full modality set or task set for each prompt.}
\label{tab:goal_parsing_prompt_model_densed}
\end{table}

On the 18-prompt intent-parsing benchmark, Table~\ref{tab:goal_parsing_prompt_model_densed} shows that \texttt{ollama-gpt-oss} achieves 88.9\% joint exact match, and \texttt{ollama-qwen3:4b} achieves 100\% joint EM, indicating that a compact 4B-parameter model can match larger backends on neuroimaging intent parsing when properly prompted. By contrast, \texttt{qwen3:14b} and \texttt{qwen3:8b} produced 100\% invalid outputs on the same prompts, indicating that intermediate model sizes in this family struggled with structured JSON generation for this task.

\begin{table*}[t]
\centering
\small
\resizebox{\textwidth}{!}{%
\begin{tabular}{lccccccc}
\toprule
Model & \# Cases & Syntax & Tool & In Path & Out Path & Step Const. & All Pass \\
\midrule
ollama-glm-4.7-flash & 33 & 100.0\% & 97.0\% & 100.0\% & 100.0\% & 78.8\% & 72.7\% \\
ollama-gpt-oss & 33 & 100.0\% & 97.0\% & 100.0\% & 100.0\% & 84.8\% & 72.7\% \\
ollama-qwen3.5:27b & 33 & 100.0\% & 97.0\% & 97.0\% & 97.0\% & 84.8\% & 84.8\% \\
ollama-qwen3.5:9b & 33 & 97.0\% & 81.8\% & 84.8\% & 84.8\% & 66.7\% & 57.6\% \\
ollama-qwen3.5:4b & 33 & 100.0\% & 78.8\% & 81.8\% & 81.8\% & 72.7\% & 60.6\% \\
ollama-qwen3.5:2b & 33 & 100.0\% & 15.2\% & 18.2\% & 18.2\% & 12.1\% & 9.1\% \\
ollama-qwen3.5:0.8b & 33 & 100.0\% & 21.2\% & 24.2\% & 24.2\% & 6.1\% & 3.0\% \\
ollama-qwen3:30b & 33 & 100.0\% & 100.0\% & 100.0\% & 100.0\% & 81.8\% & 66.7\% \\
ollama-qwen3:14b & 33 & 100.0\% & 100.0\% & 100.0\% & 100.0\% & 75.8\% & 69.7\% \\
ollama-qwen3:8b & 33 & 97.0\% & 90.9\% & 93.9\% & 87.9\% & 69.7\% & 66.7\% \\
ollama-qwen3:4b & 33 & 100.0\% & 97.0\% & 100.0\% & 100.0\% & 75.8\% & 72.7\% \\
ollama-qwen3:1.7b & 33 & 100.0\% & 54.5\% & 93.9\% & 81.8\% & 42.4\% & 0.0\% \\
\bottomrule
\end{tabular}%
}
\caption{Condensed preprocessing benchmark averaged across all modality-step combinations for each model. The reported rates are simple means over the 11 modality-step summaries in the original benchmark table, covering 33 total cases per model.}
\label{tab:preprocessing_prompt_benchmark_densed}
\end{table*}

Table~\ref{tab:preprocessing_prompt_benchmark_densed} reports preprocessing code generation quality. Syntactic correctness is high across all capable models ($\geq 97\%$). Tool correctness and path validity are also strong, but step constraint adherence (ensuring the generated code respects valid inter-step ordering and tool parameters) is the primary bottleneck: GPT-oss achieves 84.8\% and Qwen3-4b achieves 75.8\%, yielding overall ``All Pass'' rates of 72.7\% for both. The smallest model (Qwen3-1.7b) fails on every case (0\% All Pass), indicating a hard capability floor on this benchmark.

\begin{table*}[t]
\centering
\small
\resizebox{\textwidth}{!}{%
\begin{tabular}{lccccccc}
\toprule
Model & \# Cases & Row EM & Subject-Date F1 & Path Validity & Col. Completeness & Duplicate-Free & All Pass \\
\midrule
ollama-glm-4.7-flash & 8 & 50.0\% & 62.5\% & 87.5\% & 50.0\% & 100.0\% & 50.0\% \\
ollama-gpt-oss & 8 & 50.0\% & 50.0\% & 100.0\% & 50.0\% & 100.0\% & 50.0\% \\
ollama-qwen3.5:27b & 8 & 50.0\% & 62.5\% & 87.5\% & 60.7\% & 100.0\% & 50.0\% \\
ollama-qwen3.5:9b & 8 & 37.5\% & 62.5\% & 95.5\% & 37.5\% & 87.5\% & 37.5\% \\
ollama-qwen3.5:4b & 8 & 0.0\% & 62.5\% & 60.4\% & 3.4\% & 50.0\% & 0.0\% \\
ollama-qwen3.5:2b & 8 & 0.0\% & 0.0\% & 0.0\% & 0.0\% & 0.0\% & 0.0\% \\
ollama-qwen3.5:0.8b & 8 & 0.0\% & 0.0\% & 0.0\% & 0.0\% & 25.0\% & 0.0\% \\
ollama-qwen3:30b & 8 & 50.0\% & 50.0\% & 100.0\% & 50.0\% & 100.0\% & 50.0\% \\
ollama-qwen3:14b & 8 & 50.0\% & 50.0\% & 87.5\% & 50.0\% & 100.0\% & 50.0\% \\
ollama-qwen3:8b & 8 & 50.0\% & 75.0\% & 91.1\% & 50.0\% & 75.0\% & 50.0\% \\
ollama-qwen3:4b & 8 & 37.5\% & 83.3\% & 100.0\% & 46.4\% & 87.5\% & 37.5\% \\
ollama-qwen3:1.7b & 8 & 12.5\% & 25.0\% & 25.0\% & 12.5\% & 50.0\% & 12.5\% \\
\bottomrule
\end{tabular}%
}
\caption{Condensed data-integration benchmark averaged across all settings for each model. Rates are aggregated from the original setting-level summary with weighting by the number of cases in each setting, for 8 total cases per model.}
\label{tab:data_integration_benchmark_densed}
\end{table*}

Table~\ref{tab:data_integration_benchmark_densed} evaluates data integration, the stage where agents assemble multi-subject manifests from preprocessed outputs. Path validity is high ($\geq 87.5\%$ for capable models), but row-level exact match and subject-date F1 are more challenging: most models achieve 50\% row EM. The overall All Pass rate of 37.5--50\% reflects the difficulty of correctly resolving subject--session--date correspondences across modalities for open-source models, especially on multi-site heterogeneous datasets like ADNI.

\subsection{System Demonstration: Post-Interactive Group Analysis}\label{sec:post_interactive}

Beyond automated classification, NeuroAgent supports post-interactive statistical analysis via natural language queries. Unlike the downstream AD vs.\ CN classification experiments in Section~\ref{sec:classification_results}, these analyses use modality-specific matched cohorts determined by data availability and the scientific question, rather than a single unified multimodal cohort. As a demonstration, we used the agent to perform group-wise cortical-thickness analysis on FreeSurfer-derived structural MRI features and age-related regression/group-comparison analysis on preprocessed DTI biomarkers. In both cases, the agent autonomously matched subjects to available imaging visits, selected modality-appropriate features, generated statistical code, executed the analysis, produced publication-ready figures, and summarized the findings in natural language.

\subsubsection{Cortical Thickness Analysis}

We first evaluated cortical thickness using a three-group CN/MCI/AD cohort defined by the study label file. Among 2,962 labeled subjects, NeuroAgent matched 1,601 subjects to a structural MRI visit within 30 days of the reference visit (CN=651, MCI=581, AD=369; median mismatch 13.0 days). It then extracted 68 cortical thickness features from the UCSF FreeSurfer summary table and performed covariate-adjusted group analysis. Using OLS models of the form \textit{thickness} $\sim$ \textit{age} + \textit{sex} + \textit{diagnosis}, the agent recovered the expected AD-related cortical thinning pattern, with the strongest diagnosis effects in bilateral entorhinal cortex, middle temporal cortex, superior temporal cortex, inferior temporal cortex, inferior parietal cortex, fusiform cortex, supramarginal cortex, and precuneus. Bilateral entorhinal cortex remained the most discriminative cortical biomarker after multiple-comparison correction (left: $p_{\mathrm{FDR}}=1.13\times10^{-92}$; right: $p_{\mathrm{FDR}}=2.41\times10^{-85}$). Temporal and fusiform regions also showed strong diagnosis effects (e.g., left middle temporal: $p_{\mathrm{FDR}}=5.06\times10^{-87}$; left inferior temporal: $p_{\mathrm{FDR}}=6.41\times10^{-65}$; left fusiform: $p_{\mathrm{FDR}}=2.93\times10^{-58}$).

The age-stratified analysis further showed age-related thinning across all three diagnostic groups, with the steepest temporal-lobe slopes in AD and MCI. Left entorhinal thickness declined significantly with age in CN ($\beta=-0.00571$, $p=1.61\times10^{-4}$), MCI ($\beta=-0.0152$, $p=1.75\times10^{-13}$), and AD ($\beta=-0.0176$, $p=1.56\times10^{-9}$), with a similar pattern on the right. Fusiform and inferior temporal cortices showed the same graded trend, where MCI and AD both exhibited stronger negative age slopes than CN. The agent reached this output starting from a high-level request such as "analyze cortical thinning across CN, MCI, and AD" without manual scripting.

\begin{figure}[H]
    \centering
    \includegraphics[width=0.48\textwidth]{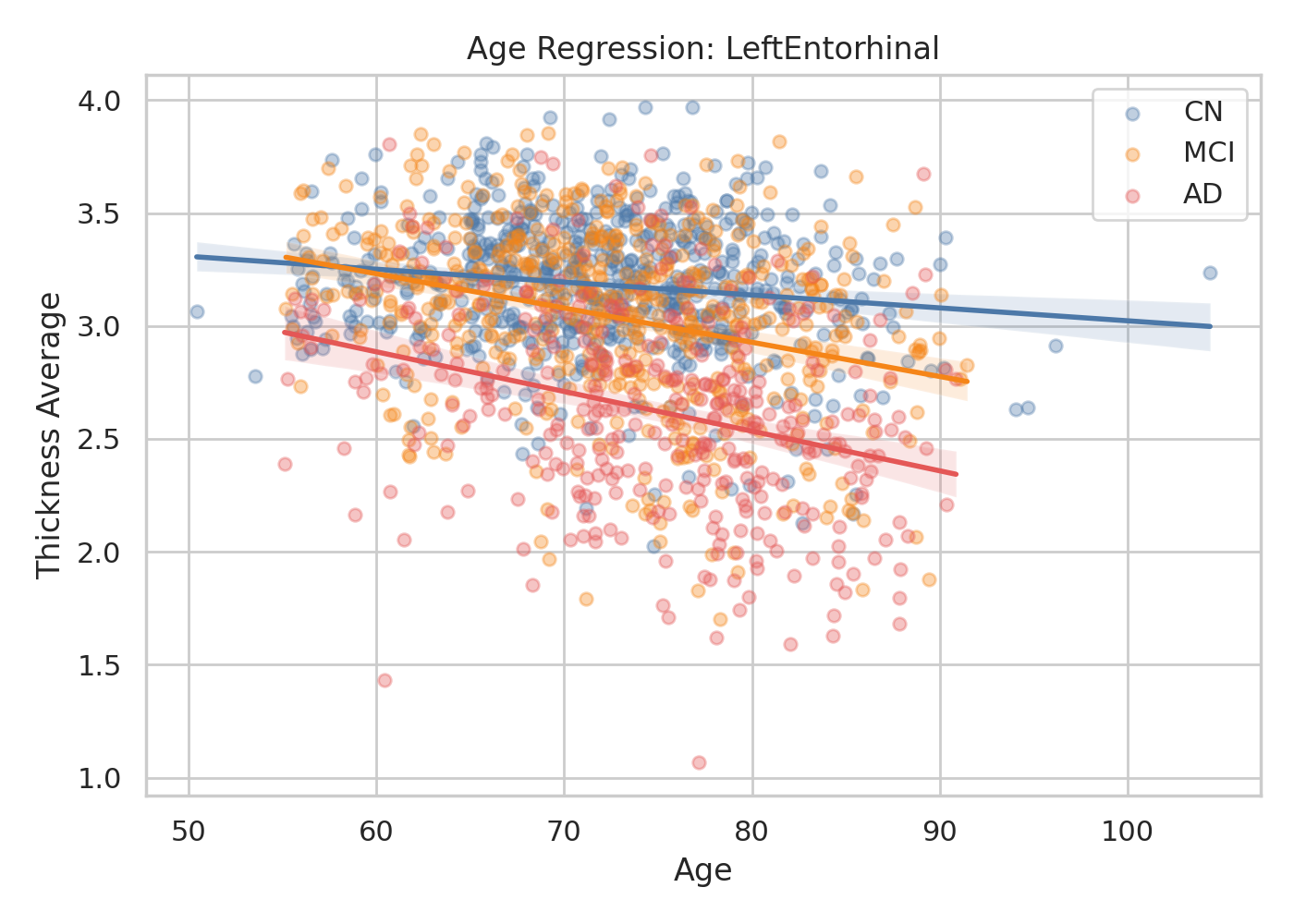}
    \hfill
    \includegraphics[width=0.48\textwidth]{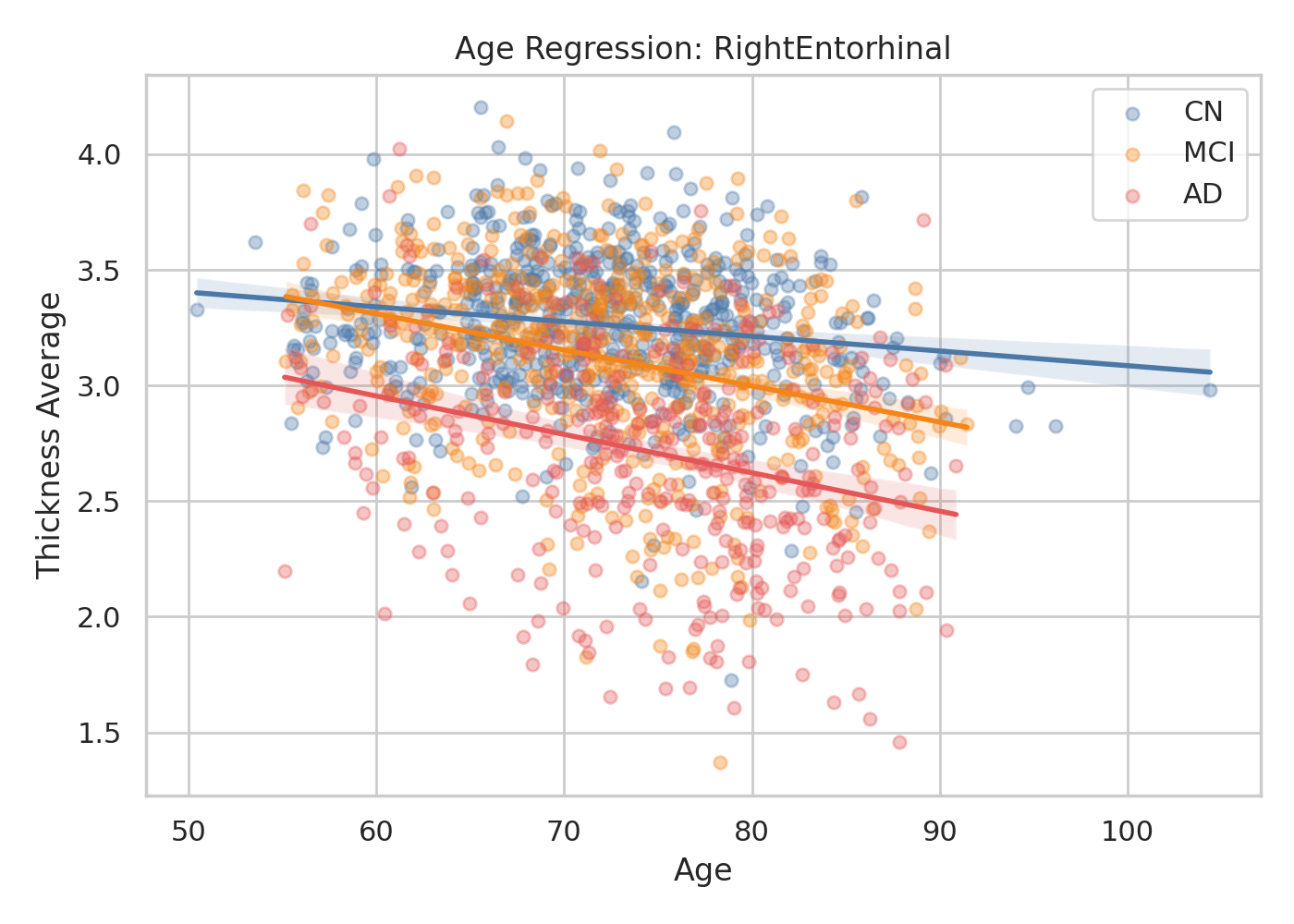}
    \caption{\textbf{Age-related Cortical Thinning in the Entorhinal Cortex (Agent-Generated).} NeuroAgent matched 1,601 labeled subjects to structural MRI visits and fit diagnosis-specific age regressions on cortical thickness. Both hemispheres show significant thinning with age, with the strongest negative slopes in MCI and AD relative to CN, consistent with progressive temporal-lobe neurodegeneration.}
    \label{fig:ct_reg_entorhinal}
\end{figure}

\begin{figure}[H]
    \centering
    \includegraphics[width=0.48\textwidth]{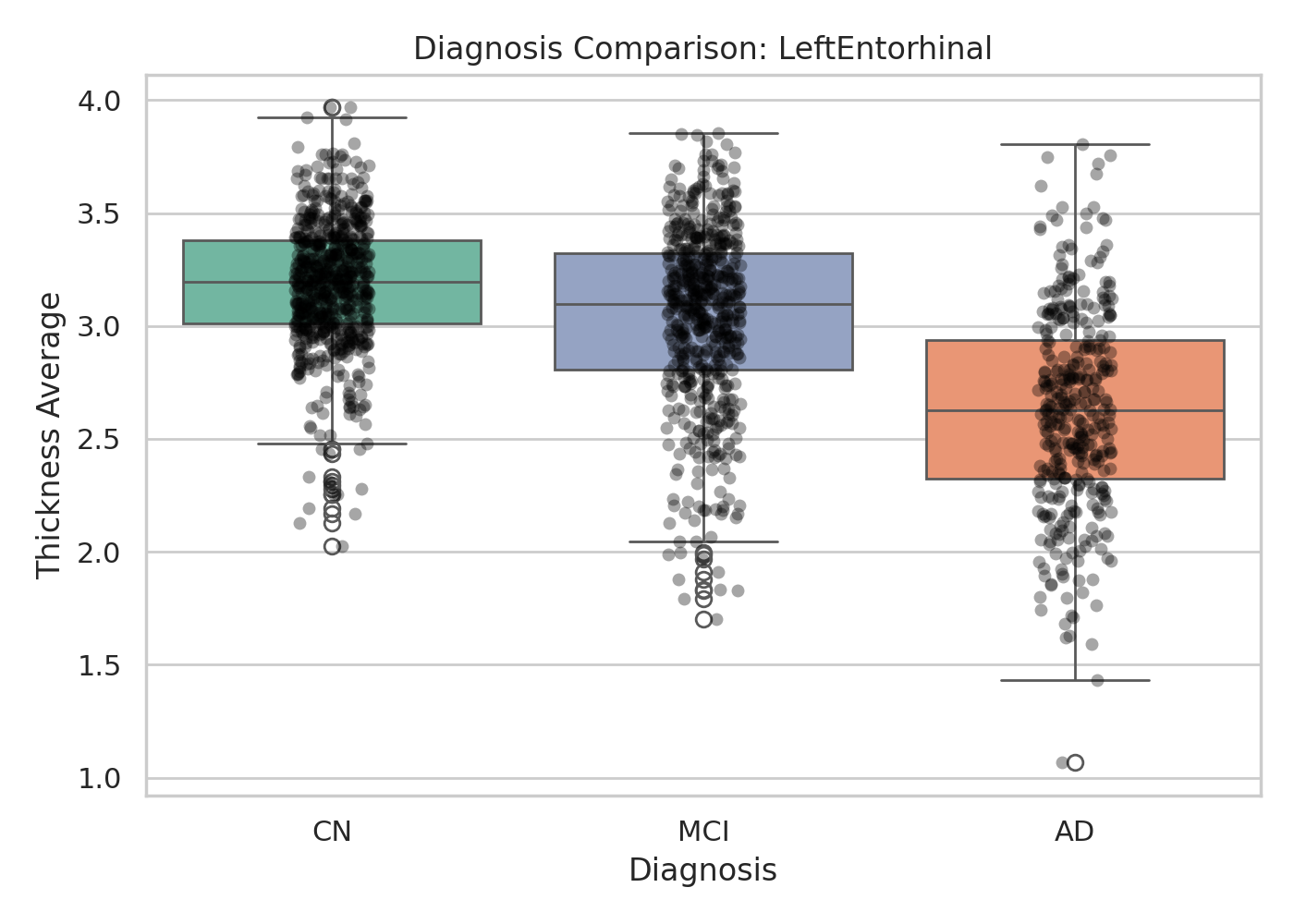}
    \hfill
    \includegraphics[width=0.48\textwidth]{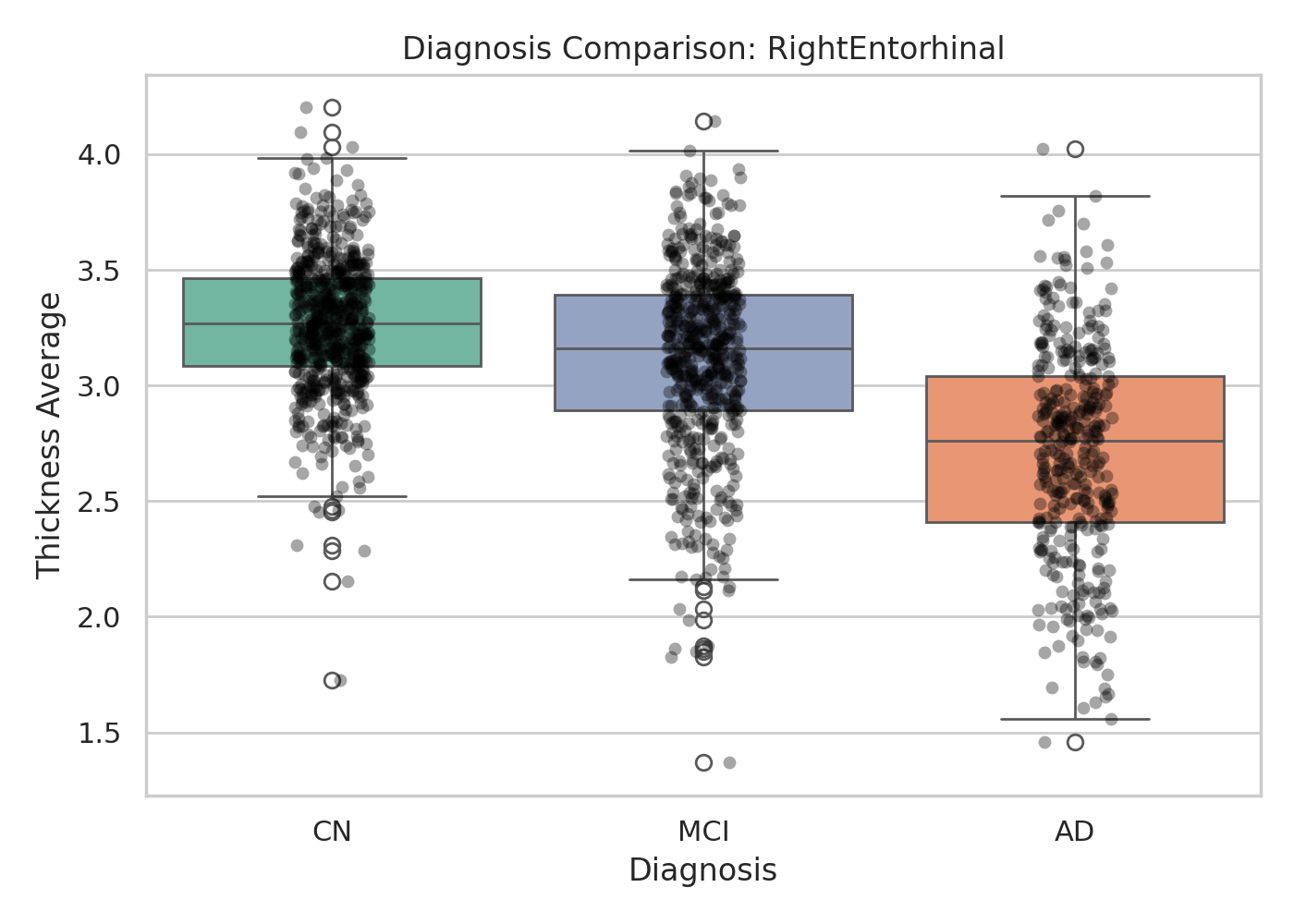}
    \caption{\textbf{Diagnosis-wise Cortical Thickness Comparison in the Entorhinal Cortex (Agent-Generated).} After matching 1,601 CN/MCI/AD subjects and adjusting for age and sex in the statistical model, NeuroAgent identified a clear stepwise reduction in entorhinal thickness from CN to MCI to AD. Bilateral entorhinal cortex ranked among the most significant cortical biomarkers in the cohort.}
    \label{fig:ct_box_entorhinal}
\end{figure}

\subsubsection{DTI Analysis}

\begin{figure}[H]
    \centering
    \includegraphics[width=0.48\textwidth]{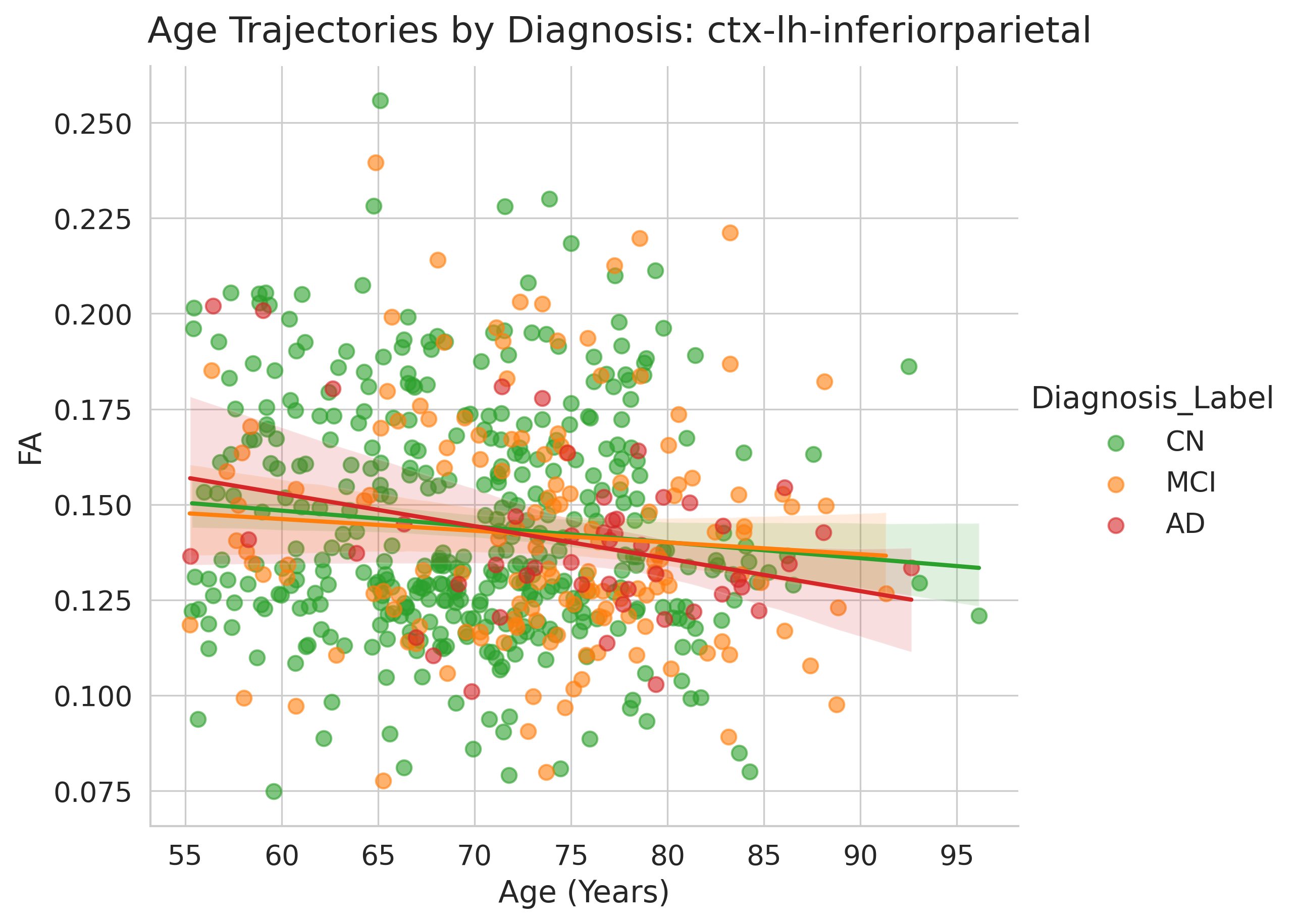}
    \hfill
    \includegraphics[width=0.48\textwidth]{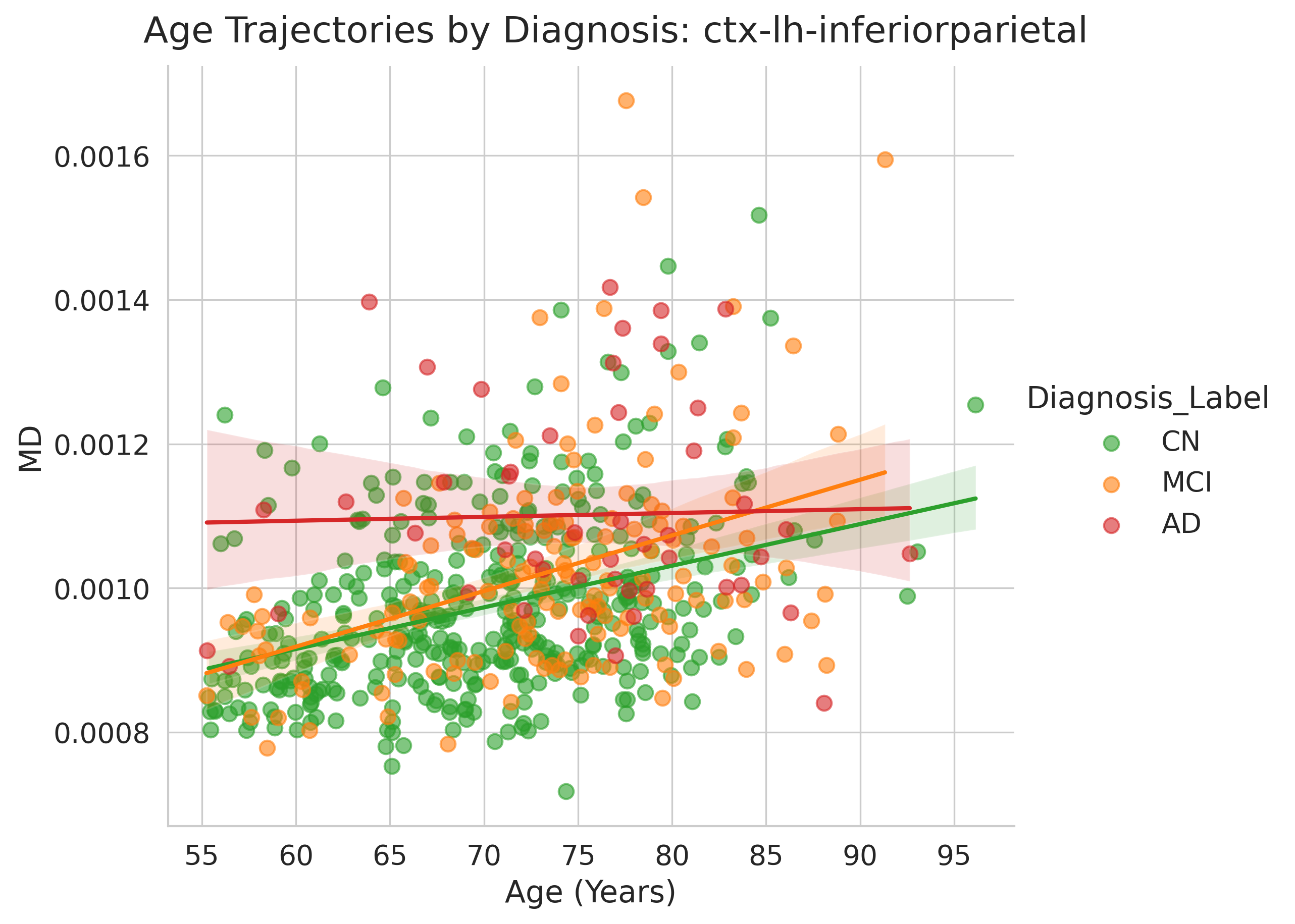}
    \caption{\textbf{Age-related Trajectories in the Inferior Parietal Cortex (Agent-Generated).} (a) FA shows a slight age-dependent decline, significant in CN ($p=0.041$) and AD ($p=0.024$) groups. (b) MD shows a robust positive correlation with age in CN ($p<0.001$) and MCI ($p<0.001$) groups, while the AD group shows no significant age effect ($p=0.842$), suggesting a ceiling effect.}
    \label{fig:reg_parietal}
\end{figure}

\begin{figure}[H]
    \centering
    \includegraphics[width=0.48\textwidth]{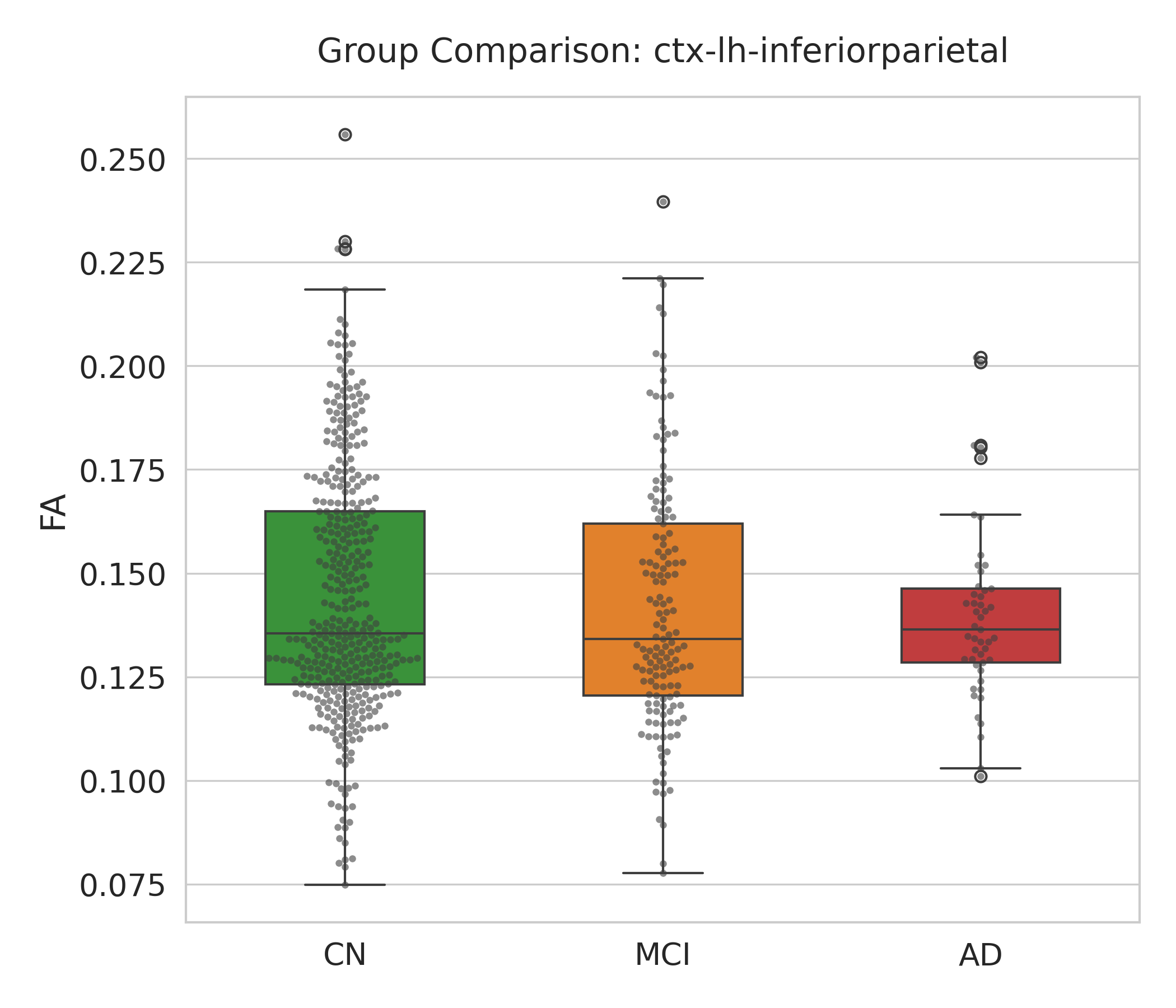}
    \hfill
    \includegraphics[width=0.48\textwidth]{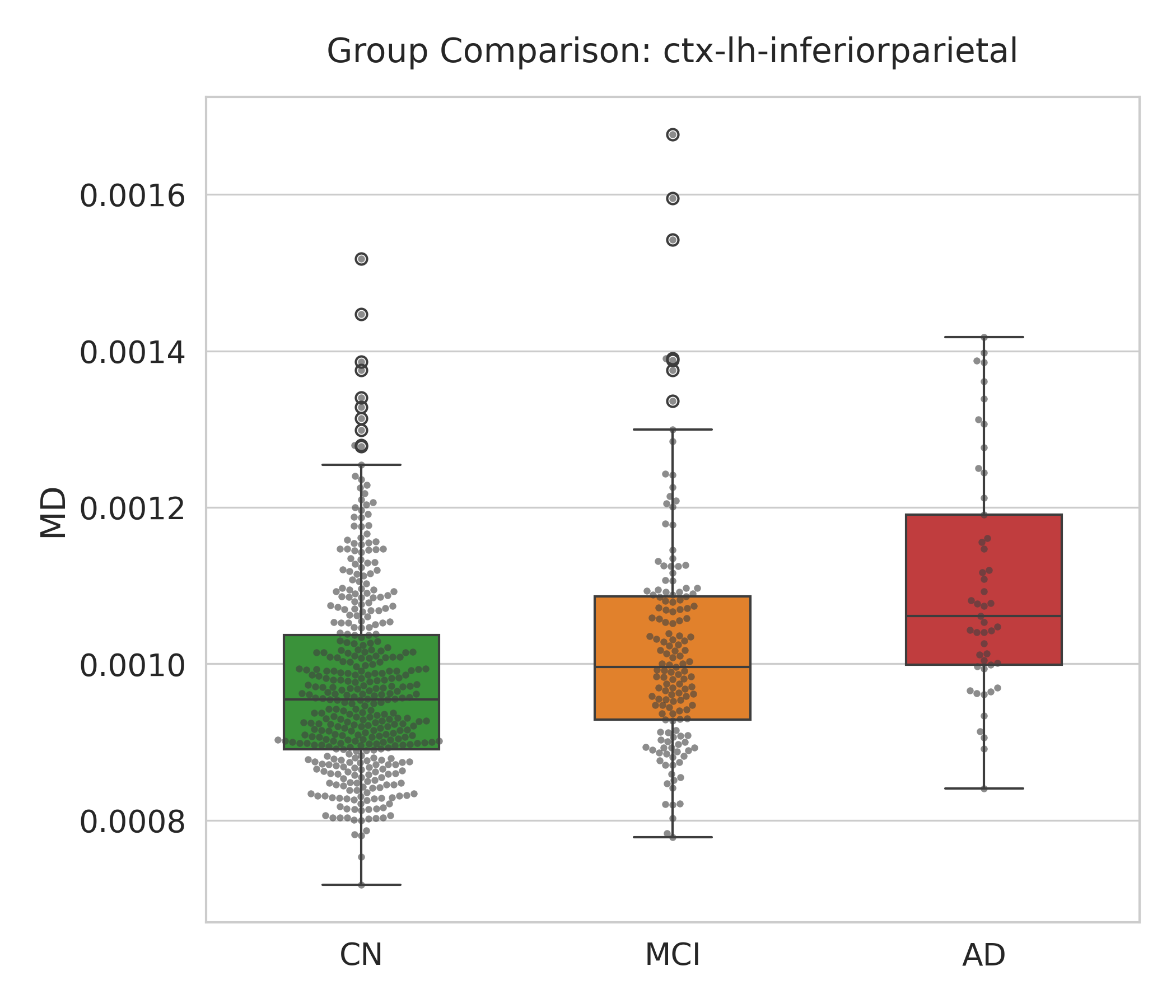}
    \caption{\textbf{Group Comparisons in the Inferior Parietal Cortex (Agent-Generated).} MD demonstrates a clear stepwise increase from CN to MCI to AD, consistent with progressive white matter degradation.}
    \label{fig:box_parietal}
\end{figure}

We next applied the same post-interactive workflow to diffusion MRI. The full DTI cohort comprises 1{,}137 unique ADNI subjects (first scan per subject; CN=707, MCI=311, AD=109, with 10 subjects missing a diagnosis label; mean age 72.9 $\pm$ 8.1 years; 56.9\% female), and is therefore broader than the AD vs.\ CN classification cohort: only 620 of these subjects (CN=563, AD=57) are nested in the 1{,}470 classification cohort reported in Table~\ref{tab:modality_availability}, while the remaining 517 are MCI cases, CN/AD subjects without a T1 in our local pool, or subjects whose diagnosis label comes only from the DTI demographic source. As a demonstration, NeuroAgent performed age-related regression and group-comparison analysis on preprocessed DTI features (FA, MD) in the inferior parietal cortex. The agent autonomously generated statistical code, executed OLS regression per diagnostic group, produced figures (Figure~\ref{fig:reg_parietal}, Figure~\ref{fig:box_parietal}), and summarized findings in natural language. The same workflow thus extends from structural MRI to diffusion MRI without manual code changes.

\subsection{AD Diagnosis Classification}\label{sec:classification_results}

We evaluated AD vs.\ CN binary classification using features automatically extracted by NeuroAgent from multiple imaging modalities. Results are reported across volumetric sMRI, Tau-PET, and multimodal combinations.

\subsubsection{Imaging and Multimodal Classification}

Tables~\ref{tab:mri_results}--\ref{tab:all_modalities_results} report AD vs.\ CN binary classification performance for sMRI, Tau-PET, sMRI+Tau-PET, and four-modality fusion (sMRI + Tau-PET + fMRI + tabular clinical), respectively. For each imaging modality configuration, we compare three 3D CNN architectures against the NeuroAgent ensemble (stratified stacking of modality-specific predictions). All results are averaged across 5-fold cross-validation with subject-level stratification. The ensemble consistently achieves the highest AUC and MCC across all settings, with the four-modality configuration reaching AUC 0.9518 and MCC 0.7389.

\begin{table}[htbp]
\centering
\caption{MRI classification performance.}
\label{tab:mri_results}
\begin{tabular}{lcccccc}
\hline
Method & Accuracy & Precision & Recall & F1-Score & AUC & MCC \\
\hline
ResNet-18 & 0.7333 & 0.5855 & 0.5681 & 0.5767 & 0.7815 & 0.3822 \\
DenseNet-121 & 0.5374 & 0.3977 & 0.8681 & 0.5455 & 0.6293 & 0.2540 \\
EfficientNet-B0 & 0.7313 & 0.5728 & 0.6277 & 0.5990 & 0.7735 & 0.3985 \\
Agent Ensemble & 0.8293 & 0.7786 & 0.6511 & 0.7092 & 0.8624 & 0.5944 \\
\hline
\end{tabular}
\end{table}

\begin{table}[htbp]
\centering
\caption{Tau-PET classification performance.}
\label{tab:taupet_results}
\begin{tabular}{lcccccc}
\hline
Method & Accuracy & Precision & Recall & F1-Score & AUC & MCC \\
\hline
ResNet-18 & 0.9190 & 0.7660 & 0.5714 & 0.6545 & 0.9096 & 0.6182 \\
DenseNet-121 & 0.8913 & 0.5698 & 0.7778 & 0.6577 & 0.9159 & 0.6051 \\
EfficientNet-B0 & 0.8252 & 0.3908 & 0.5397 & 0.4533 & 0.7510 & 0.3589 \\
Agent Ensemble & 0.9147 & 0.7556 & 0.5397 & 0.6296 & 0.9092 & 0.5935 \\
\hline
\end{tabular}
\end{table}

\begin{table}[htbp]
\centering
\caption{Tau-PET and MRI multimodal classification performance with per-model multimodal baselines and stratified stacking.}
\label{tab:taupet_mri_results}
\begin{tabular}{lcccccc}
\hline
Method & Accuracy & Precision & Recall & F1-Score & AUC & MCC \\
\hline
ResNet-18 & 0.7776 & 0.6765 & 0.5830 & 0.6263 & 0.8455 & 0.4718 \\
DenseNet-121 & 0.5946 & 0.4345 & 0.8894 & 0.5838 & 0.6945 & 0.3387 \\
EfficientNet-B0 & 0.7646 & 0.6330 & 0.6277 & 0.6303 & 0.8090 & 0.4577 \\
Agent Ensemble & 0.8565 & 0.8246 & 0.7000 & 0.7572 & 0.9117 & 0.6607 \\
\hline
\end{tabular}
\end{table}

\begin{table}[htbp]
\centering
\caption{Four-modality classification performance (sMRI + Tau-PET + fMRI + tabular clinical) with per-model baselines and stratified stacking.}
\label{tab:all_modalities_results}
\begin{tabular}{lcccccc}
\hline
Method & Accuracy & Precision & Recall & F1-Score & AUC & MCC \\
\hline
ResNet-18 & 0.8333 & 0.8032 & 0.6340 & 0.7087 & 0.9149 & 0.6024 \\
DenseNet-121 & 0.7156 & 0.5327 & 0.9000 & 0.6693 & 0.8270 & 0.4950 \\
EfficientNet-B0 & 0.8551 & 0.8142 & 0.7085 & 0.7577 & 0.9059 & 0.6583 \\
Agent Ensemble & 0.8878 & 0.8427 & 0.7979 & 0.8197 & 0.9518 & 0.7389 \\
\hline
\end{tabular}
\end{table}

EfficientNet-B0 underperforms on Tau-PET (F1 0.4533, AUC 0.7510) due to per-fold instability rather than poor overall ranking quality: per-fold AUC varies from 0.634 to 0.934 across the five folds, and in one fold the model collapses to predicting CN for every test subject. This fragility likely stems from training the smallest backbone (4.7M parameters), whose BatchNorm-heavy MBConv architecture is sensitive to small-batch statistics, on the relatively small and imbalanced Tau-PET cohort (n=469, AD prevalence 13.4\%). The Agent Ensemble nonetheless reaches AUC 0.9092 on Tau-PET, on par with ResNet-18 (AUC 0.9096) and DenseNet-121 (AUC 0.9159), indicating that the MLP stacker absorbs the noisier EfficientNet-B0 feature without degradation.

\section{Discussion}\label{sec:discussion}

Our ablation results indicate that LLM backend capability is a decisive factor in pipeline reliability, but is not always monotone in parameter count. \texttt{qwen3:4b} matches or exceeds \texttt{qwen3:14b} and \texttt{qwen3:8b} on both intent parsing and preprocessing benchmarks. A likely cause is that smaller models in this family default to structured output modes more reliably, whereas larger ones produce verbose chain-of-thought that breaks JSON parsing. This finding has practical implications: researchers without access to large proprietary models can use a compact, locally deployable backend without sacrificing pipeline accuracy on the benchmarks reported here. Step-constraint adherence (72.7--75.8\% All Pass for the best models) remains the principal bottleneck; the 27--30\% failure rate means that roughly 1 in 4 preprocessing cases requires retry or HITL escalation. The current benchmarks (18 parsing prompts, 33 preprocessing cases, 8 integration cases) are sufficient to reveal systematic differences across backends and to expose the present limits of open-source models on structured-output neuroimaging tasks, but are not large enough to precisely characterize tail failure rates or to generalize to all modalities and edge cases. Expanding these benchmarks is therefore a clear next step.

Beyond pipeline-level reliability, the agent ensemble outperforms individual models across all modality configurations on the downstream classification task. Four-modality fusion (sMRI + Tau-PET + fMRI + tabular clinical) achieves AUC 0.9518, compared to AUC 0.8624 for sMRI-only and AUC 0.9096 for Tau-PET-only baselines, indicating that the automated preprocessing pipeline preserves enough diagnostic signal across modalities to support multimodal fusion. The gap between the agent ensemble and single-architecture baselines (e.g., ResNet-18 at AUC 0.9149 with four modalities) is also consistent with the stratified stacking approach. We do not compare against a conventional manually-scripted preprocessing pipeline in this version; future work should include such a comparison to directly quantify any quality difference attributable to the automated approach.

The value of the system extends past preprocessing and classification. The cortical-thickness and DTI demonstrations show that, once data is structured by the pipeline, users can pose statistical questions in natural language and receive analysis outputs without writing any code. This last-mile capability, from preprocessed data to scientific figures, is a feature not provided by traditional workflow managers such as Nipype or fMRIPrep.

Several limitations qualify these results. The current version focuses on AD classification as the downstream task; regression and segmentation evaluations are left for future work. Pipeline latency is also non-trivial: full preprocessing for a single subject takes 4--8 hours depending on modality and cluster availability, with additional LLM API overhead per step. The reflective error recovery, while effective for common failure modes (syntax errors, path mismatches), may not handle rare domain-specific failures (e.g., extreme image artifacts) without human escalation via the HITL interface, so a natural next step is to integrate modality-specific quality-control tools. The current benchmark success rates of 72.7\% (preprocessing) and 37.5--50\% (data integration) further suggest that meaningful HITL involvement is still required for a non-trivial fraction of cases, especially when relying on open-source backends. The pipeline-ablation benchmarks themselves are intentionally small-scale for this first-version preprint: the current sizes (18 parsing prompts, 33 preprocessing cases, 8 integration cases) establish a clear evaluation framework but are not sufficient to claim general robustness. A comparison against conventional manually-scripted preprocessing pipelines on downstream classification quality is also absent and will be addressed in future work. Finally, the ADNI cohort, though large, has relatively standardized acquisition protocols; generalization to real-world heterogeneous clinical datasets requires separate evaluation.

\section{Conclusion}\label{sec:conclusion}

We presented NeuroAgent, an LLM agent framework for autonomous multimodal neuroimaging preprocessing and analysis. By combining hierarchical multi-agent orchestration, a feedback-driven Generate-Execute-Validate execution engine, and an interactive HITL interface, NeuroAgent automates the complete lifecycle from raw imaging data to scientific results without manual pipeline configuration. Our evaluation on 1,470 ADNI subjects (CN=1,000, AD=470) shows substantial pipeline automation: capable LLM backends achieve up to 100\% intent-parsing accuracy and 72.7\% end-to-end preprocessing correctness, with a retrying loop and an HITL interface handling the remaining failure cases. Data integration (37.5--50\% all-pass) is the main bottleneck for future improvement. Once data are structured by the pipeline, NeuroAgent supports natural-language-driven statistical analysis and visualization: in our cortical-thickness and DTI demonstrations the agent reproduced clinically expected effects (AD-related cortical thinning and age-related FA decline) directly from a high-level natural-language request, letting clinical researchers obtain analysis outputs from preprocessed data without programming expertise. Building on the same automatically preprocessed data, the multimodal agent ensemble further achieves AUC 0.9518 across four modalities for AD vs.\ CN classification, indicating that the pipeline preserves enough diagnostic information for downstream learning. Together, these results position NeuroAgent as a step toward more autonomous scientific workflows in computational neuroscience, where less manual effort sits between data acquisition and scientific conclusions.

\section*{Acknowledgment}
This work was supported by the National Institute of Health (NIH) under grants R01EB022744, RF1AG077578,  R01AG064584, U19AG078109, and P30AG066530.

Data used in preparation of this article were obtained from the Alzheimer’s Disease Neuroimaging Initiative (ADNI) database (adni.loni.usc.edu). As such, the investigators within the ADNI contributed to the design and implementation of ADNI and/or provided data but did not participate in analysis or writing of this report. A complete listing of ADNI investigators can be found at: \url{http://adni.loni.usc.edu/wp-content/uploads/how_to_apply/ADNI_Acknowledgement_List.pdf}


\bibliographystyle{references}
\bibliography{sn-bibliography}

\begin{thebibliography}{41}
\providecommand{\natexlab}[1]{#1}
\providecommand{\url}[1]{\texttt{#1}}
\expandafter\ifx\csname urlstyle\endcsname\relax
  \providecommand{\doi}[1]{doi: #1}\else
  \providecommand{\doi}{doi: \begingroup \urlstyle{rm}\Url}\fi

\bibitem[Arani et~al.(2024)Arani, Borowski, Felmlee, Reid, Thomas, Gunter, Stables, Buckner, Jung, Tosun, et~al.]{arani2024design}
Arvin Arani, Bret Borowski, John Felmlee, Robert~I Reid, David~L Thomas, Jeffrey~L Gunter, Lara Stables, Randy~L Buckner, Youngkyoo Jung, Duygu Tosun, et~al.
\newblock Design and validation of the adni mr protocol.
\newblock \emph{Alzheimer's \& Dementia}, 20\penalty0 (9):\penalty0 6615--6621, 2024.

\bibitem[Arani et~al.(2025)Arani, Bernstein, Borowski, Jack, and Weiner]{arani2025adni4}
Arvin Arani, Matthew~A. Bernstein, Brian~J. Borowski, Clifford~R. Jack, and Michael~W. Weiner.
\newblock Design and validation of the adni4 mri protocol.
\newblock \emph{Alzheimer's \& Dementia}, 2025.
\newblock In press.

\bibitem[Ashburner \& Friston(2005)Ashburner and Friston]{ashburner2005spm}
John Ashburner and Karl~J. Friston.
\newblock Unified segmentation.
\newblock \emph{NeuroImage}, 26\penalty0 (3):\penalty0 839--851, 2005.
\newblock \doi{10.1016/j.neuroimage.2005.02.018}.

\bibitem[Avants et~al.(2011)Avants, Tustison, Song, Cook, Klein, and Gee]{avants2011ants}
Brian~B. Avants, Nicholas~J. Tustison, Gang Song, Philip~A. Cook, Arno Klein, and James~C. Gee.
\newblock A reproducible evaluation of {ANTs} similarity metric performance in brain image registration.
\newblock \emph{NeuroImage}, 54\penalty0 (3):\penalty0 2033--2044, 2011.
\newblock \doi{10.1016/j.neuroimage.2010.09.025}.

\bibitem[Bai et~al.(2024)Bai, Li, Ling, Kim, and Zhao]{bai2024sparsellm}
Guangji Bai, Yijiang Li, Chen Ling, Kibaek Kim, and Liang Zhao.
\newblock Sparsellm: Towards global pruning for pre-trained language models, 2024.
\newblock URL \url{https://arxiv.org/abs/2402.17946}.

\bibitem[Basser et~al.(1994)Basser, Mattiello, and LeBihan]{basser1994dti}
Peter~J. Basser, James Mattiello, and Denis LeBihan.
\newblock Mr diffusion tensor spectroscopy and imaging.
\newblock \emph{Journal of Magnetic Resonance, Series B}, 103\penalty0 (3):\penalty0 247--254, 1994.
\newblock \doi{10.1006/jmrb.1994.1037}.

\bibitem[Botvinik-Nezer et~al.(2020)Botvinik-Nezer, Holzmeister, Camerer, Dreber, Huber, Johannesson, Kirchler, Iwanir, Mumford, Adcock, et~al.]{botviniknezer2020variability}
Rotem Botvinik-Nezer, Felix Holzmeister, Colin~F. Camerer, Anna Dreber, Juergen Huber, Magnus Johannesson, Michael Kirchler, Roni Iwanir, Jeanette~A. Mumford, R.~Alison Adcock, et~al.
\newblock Variability in the analysis of a single neuroimaging dataset by many teams.
\newblock \emph{Nature}, 582\penalty0 (7810):\penalty0 84--88, 2020.
\newblock \doi{10.1038/s41586-020-2314-9}.
\newblock URL \url{https://doi.org/10.1038/s41586-020-2314-9}.

\bibitem[Cardoso et~al.(2022)Cardoso, Li, Brown, Ma, Kerfoot, Wang, Murrey, Myronenko, Zhao, Yang, et~al.]{cardoso2022monai}
M~Jorge Cardoso, Wenqi Li, Richard Brown, Nic Ma, Eric Kerfoot, Yiheng Wang, Benjamin Murrey, Andriy Myronenko, Can Zhao, Dong Yang, et~al.
\newblock Monai: An open-source framework for deep learning in healthcare.
\newblock \emph{arXiv preprint arXiv:2211.02701}, 2022.

\bibitem[Chen et~al.(2026)]{chen2026adreasoning}
Qiuhui Chen et~al.
\newblock Ad-reasoning: Multimodal guideline-guided reasoning for alzheimer's disease diagnosis, 2026.
\newblock URL \url{https://arxiv.org/abs/2603.24059}.

\bibitem[Esteban et~al.(2019{\natexlab{a}})Esteban, Blair, Nielson, Varada, Marrett, Thomas, Poldrack, and Gorgolewski]{esteban2019mriqc}
Oscar Esteban, Ross~W. Blair, Dylan~M. Nielson, Jan~C. Varada, Sean Marrett, Adam~G. Thomas, Russell~A. Poldrack, and Krzysztof~J. Gorgolewski.
\newblock Crowdsourced {MRI} quality metrics and expert quality annotations for training of humans and machines.
\newblock \emph{Scientific Data}, 6\penalty0 (1):\penalty0 30, 2019{\natexlab{a}}.
\newblock \doi{10.1038/s41597-019-0035-4}.

\bibitem[Esteban et~al.(2019{\natexlab{b}})]{esteban2019fmriprep}
Oscar Esteban et~al.
\newblock fmriprep: a robust preprocessing pipeline for functional mri.
\newblock \emph{Nature methods}, 16\penalty0 (1):\penalty0 111--116, 2019{\natexlab{b}}.

\bibitem[Fan et~al.(2026)]{fan2026macro}
Lin Fan et~al.
\newblock Evolving medical imaging agents via experience-driven self-skill discovery, 2026.
\newblock URL \url{https://arxiv.org/abs/2603.05860}.

\bibitem[Fathi et~al.(2025)Fathi, Kumar, and Arbel]{fathi2025aura}
Nima Fathi, Amar Kumar, and Tal Arbel.
\newblock Aura: A multi-modal medical agent for understanding, reasoning \& annotation, 2025.
\newblock URL \url{https://arxiv.org/abs/2507.16940}.

\bibitem[Fischl(2012)]{fischl2012freesurfer}
Bruce Fischl.
\newblock {FreeSurfer}.
\newblock \emph{NeuroImage}, 62\penalty0 (2):\penalty0 774--781, 2012.
\newblock \doi{10.1016/j.neuroimage.2012.01.021}.

\bibitem[Gorgolewski et~al.(2011)Gorgolewski, Burns, Madison, Clark, Halchenko, Waskom, and Ghosh]{gorgolewski2011nipype}
Krzysztof Gorgolewski, Christopher~D. Burns, Cindee Madison, Dav Clark, Yaroslav~O. Halchenko, Michael~L. Waskom, and Satrajit~S. Ghosh.
\newblock Nipype: A flexible, lightweight and extensible neuroimaging data processing framework in python.
\newblock \emph{Frontiers in Neuroinformatics}, 5, 2011.
\newblock \doi{10.3389/fninf.2011.00013}.
\newblock URL \url{https://doi.org/10.3389/fninf.2011.00013}.

\bibitem[Hou et~al.(2025)Hou, Yang, Du, Lau, Liu, He, Long, and Wang]{hou2025adagent}
Wenlong Hou, Guangqian Yang, Ye~Du, Yeung Lau, Lihao Liu, Junjun He, Ling Long, and Shujun Wang.
\newblock Adagent: Llm agent for alzheimer's disease analysis with collaborative coordinator, 2025.
\newblock URL \url{https://arxiv.org/abs/2506.11150}.

\bibitem[Hou et~al.(2026)Hou, Bi, Yang, Liu, Du, Xue, Wang, Feng, Xun, Yu, et~al.]{hou2026adcare}
Wenlong Hou, Sheng Bi, Guangqian Yang, Lihao Liu, Ye~Du, Hanxiao Xue, Juncheng Wang, Yuxiang Feng, Yue Xun, Nanxi Yu, et~al.
\newblock Ad-care: A guideline-grounded, modality-agnostic llm agent for real-world alzheimer's disease diagnosis with multi-cohort assessment, fairness analysis, and reader study, 2026.
\newblock URL \url{https://arxiv.org/abs/2603.25322}.

\bibitem[Jack et~al.(2018)Jack, Bennett, Blennow, Carrillo, Dunn, Haeberlein, Holtzman, Jagust, Jessen, Karlawish, et~al.]{jack2018niaaa}
Clifford~R. Jack, David~A. Bennett, Kaj Blennow, Maria~C. Carrillo, Billy Dunn, Samantha~Budd Haeberlein, David~M. Holtzman, William Jagust, Frank Jessen, Jason Karlawish, et~al.
\newblock Nia-aa research framework: Toward a biological definition of alzheimer's disease.
\newblock \emph{Alzheimer's \& Dementia}, 14\penalty0 (4):\penalty0 535--562, 2018.
\newblock \doi{10.1016/j.jalz.2018.02.018}.

\bibitem[Jack et~al.(2024)Jack, Andrews, Beach, Buracchio, Dunn, Graf, Hansson, Ho, Jagust, McDade, et~al.]{jack2024revised}
Clifford~R. Jack, J.~Scott Andrews, Thomas~G. Beach, Teresa Buracchio, Billy Dunn, Ana Graf, Oskar Hansson, Carole Ho, William Jagust, Eric McDade, et~al.
\newblock Revised criteria for diagnosis and staging of alzheimer's disease: Alzheimer's association workgroup.
\newblock \emph{Alzheimer's \& Dementia}, 20\penalty0 (8):\penalty0 5143--5169, 2024.
\newblock \doi{10.1002/alz.13859}.

\bibitem[Jenkinson et~al.(2012)Jenkinson, Beckmann, Behrens, Woolrich, and Smith]{jenkinson2012fsl}
Mark Jenkinson, Christian~F Beckmann, Timothy~EJ Behrens, Mark~W Woolrich, and Stephen~M Smith.
\newblock Fsl.
\newblock \emph{Neuroimage}, 62\penalty0 (2):\penalty0 782--790, 2012.

\bibitem[Kenia et~al.(2025)]{kenia2025rexmle}
Roshan Kenia et~al.
\newblock Rex-mle: The autonomous agent benchmark for medical imaging challenges, 2025.
\newblock URL \url{https://arxiv.org/abs/2512.17838}.

\bibitem[Klein et~al.(2010)Klein, Staring, Murphy, Viergever, and Pluim]{klein2010elastix}
Stefan Klein, Marius Staring, Keelin Murphy, Max~A. Viergever, and Josien P.~W. Pluim.
\newblock elastix: A toolbox for intensity-based medical image registration.
\newblock \emph{IEEE Transactions on Medical Imaging}, 29\penalty0 (1):\penalty0 196--205, 2010.
\newblock \doi{10.1109/TMI.2009.2035616}.

\bibitem[Li et~al.(2025)Li, Xu, Bao, Liu, Liu, Liu, Wang, Lei, Wang, Xu, Cui, Yao, Koga, and Huang]{li2025tissuelab}
Songhao Li, Jonathan Xu, Tiancheng Bao, Yuxuan Liu, Yuchen Liu, Yihang Liu, Lilin Wang, Wenhui Lei, Sheng Wang, Yinuo Xu, Yan Cui, Jialu Yao, Shunsuke Koga, and Zhi Huang.
\newblock A co-evolving agentic ai system for medical imaging analysis, 2025.
\newblock URL \url{https://arxiv.org/abs/2509.20279}.

\bibitem[Loftus et~al.(2023)Loftus, Puri, and Meyers]{loftus2023multimodality}
James~Ryan Loftus, Savita Puri, and Steven~P. Meyers.
\newblock Multimodality imaging of neurodegenerative disorders with a focus on multiparametric magnetic resonance and molecular imaging.
\newblock \emph{Insights into Imaging}, 14\penalty0 (1), 2023.
\newblock \doi{10.1186/s13244-022-01358-6}.

\bibitem[Lundberg \& Lee(2017)Lundberg and Lee]{lundberg2017unified}
Scott~M. Lundberg and Su-In Lee.
\newblock A unified approach to interpreting model predictions.
\newblock In \emph{Advances in Neural Information Processing Systems (NeurIPS)}, pp.\  4765--4774, 2017.

\bibitem[Luo et~al.(2024)Luo, Rechardt, Sun, Nejad, Y{\'a}{\~n}ez, Yilmaz, Lee, Cohen, Borghesani, Pashkov, et~al.]{luo2024brainbench}
Xiaoliang Luo, Akilles Rechardt, Guangzhi Sun, Kevin~K. Nejad, Felipe Y{\'a}{\~n}ez, Bati Yilmaz, Kangjoo Lee, Alexandra~O. Cohen, Valentina Borghesani, Anton Pashkov, et~al.
\newblock Large language models surpass human experts in predicting neuroscience results.
\newblock \emph{Nature Human Behaviour}, 9\penalty0 (2):\penalty0 305--315, nov 2024.
\newblock \doi{10.1038/s41562-024-02046-9}.
\newblock URL \url{https://doi.org/10.1038/s41562-024-02046-9}.

\bibitem[Odusami et~al.(2023)Odusami, Maskeli{\=u}nas, Dama{\v{s}}evi{\v{c}}ius, and Misra]{odusami2023machine}
Modupe Odusami, Rytis Maskeli{\=u}nas, Robertas Dama{\v{s}}evi{\v{c}}ius, and Sanjay Misra.
\newblock Machine learning with multimodal neuroimaging data to classify stages of alzheimer's disease: a systematic review and meta-analysis.
\newblock \emph{Cognitive Neurodynamics}, 18\penalty0 (3):\penalty0 775--794, 2023.
\newblock \doi{10.1007/s11571-023-09993-5}.

\bibitem[Power et~al.(2012)Power, Barnes, Snyder, Schlaggar, and Petersen]{power2012motion}
Jonathan~D. Power, Kelly~A. Barnes, Abraham~Z. Snyder, Bradley~L. Schlaggar, and Steven~E. Petersen.
\newblock Spurious but systematic correlations in functional connectivity {MRI} networks arise from subject motion.
\newblock \emph{NeuroImage}, 59\penalty0 (3):\penalty0 2142--2154, 2012.
\newblock \doi{10.1016/j.neuroimage.2011.10.018}.

\bibitem[Qian et~al.(2026)Qian, Hu, Yu, Xin, Chen, Zhang, Jiang, Liu, and Li]{qian2026medmaslab}
Yunhang Qian, Xiaobin Hu, Jiaquan Yu, Siyang Xin, Xiaokun Chen, Jiangning Zhang, Peng-Tao Jiang, Jiawei Liu, and Hongwei~Bran Li.
\newblock Medmaslab: A unified orchestration framework for benchmarking multimodal medical multi-agent systems, 2026.
\newblock URL \url{https://arxiv.org/abs/2603.09909}.

\bibitem[Sajua et~al.(2025)Sajua, Akhib, and Chang]{sajua2025agentmri}
Gulfam~Ahmed Sajua, Marjan Akhib, and Yuchou Chang.
\newblock Agentmri: A vison language model-powered ai system for self-regulating mri reconstruction with multiple degradations.
\newblock \emph{Journal of Imaging Informatics in Medicine}, jul 2025.
\newblock \doi{10.1007/s10278-025-01617-0}.
\newblock URL \url{https://doi.org/10.1007/s10278-025-01617-0}.

\bibitem[Schick et~al.(2023)Schick, Dwivedi-Yu, Dess{\`i}, Raileanu, Lomeli, Zettlemoyer, Cancedda, and Scialom]{schick2023toolformer}
Timo Schick, Jane Dwivedi-Yu, Roberto Dess{\`i}, Roberta Raileanu, Maria Lomeli, Luke Zettlemoyer, Nicola Cancedda, and Thomas Scialom.
\newblock Toolformer: Language models can teach themselves to use tools, 2023.
\newblock URL \url{https://arxiv.org/abs/2302.04761}.

\bibitem[Sellergren et~al.(2025)Sellergren, Kazemzadeh, Jaroensri, Kiraly, Traverse, Kohlberger, Xu, Jamil, Hughes, Lau, et~al.]{sellergren2025medgemma}
Andrew Sellergren, Sahar Kazemzadeh, Tiam Jaroensri, Atilla Kiraly, Madeleine Traverse, Timo Kohlberger, Shawn Xu, Fayaz Jamil, C{\'i}an Hughes, Charles Lau, et~al.
\newblock Medgemma technical report, 2025.
\newblock URL \url{https://arxiv.org/abs/2507.05201}.

\bibitem[Shen et~al.(2017)Shen, Wu, and Suk]{shen2017deep}
Dinggang Shen, Guorong Wu, and Heung-Il Suk.
\newblock Deep learning in medical image analysis.
\newblock \emph{Annual review of biomedical engineering}, 19:\penalty0 221--248, 2017.

\bibitem[Singhal et~al.(2023)Singhal, Tu, Gottweis, Sayres, Wulczyn, Hou, Clark, Pfohl, Cole-Lewis, Neal, et~al.]{singhal2023medpalm}
Karan Singhal, Tao Tu, Juraj Gottweis, Rory Sayres, Ellery Wulczyn, Le~Hou, Kevin Clark, Stephen Pfohl, Heather Cole-Lewis, Darlene Neal, et~al.
\newblock Towards expert-level medical question answering with large language models, 2023.
\newblock URL \url{https://arxiv.org/abs/2305.09617}.

\bibitem[Smith(2002)]{smith2002bet}
Stephen~M. Smith.
\newblock Fast robust automated brain extraction.
\newblock \emph{Human Brain Mapping}, 17\penalty0 (3):\penalty0 143--155, 2002.
\newblock \doi{10.1002/hbm.10062}.

\bibitem[Smith et~al.(2013)]{smith2013resting}
Stephen~M Smith et~al.
\newblock Resting-state fmri in the human connectome project.
\newblock \emph{Neuroimage}, 80:\penalty0 144--168, 2013.

\bibitem[Sozer et~al.(2025)Sozer, Sahin, Sozer, Erol, Tufek, Nernekli, Demirtas, and Celtikci]{sozer2025mri}
Alperen Sozer, Mustafa~Caglar Sahin, Batuhan Sozer, Gokberk Erol, Ozan~Yavuz Tufek, Kerem Nernekli, Zuhal Demirtas, and Emrah Celtikci.
\newblock Do llms have 'the eye' for mri? evaluating gpt-4o, grok, and gemini on brain mri performance: First evaluation of grok in medical imaging and a comparative analysis.
\newblock \emph{Diagnostics}, 15\penalty0 (11):\penalty0 1320, may 2025.
\newblock \doi{10.3390/diagnostics15111320}.
\newblock URL \url{https://doi.org/10.3390/diagnostics15111320}.

\bibitem[Wang et~al.(2024)]{wang2024survey}
Lei Wang et~al.
\newblock A survey on large language model based autonomous agents.
\newblock \emph{Frontiers of Computer Science}, 2024.

\bibitem[Wu et~al.(2023)Wu, Bansal, Zhang, Wu, Li, Zhu, Jiang, Zhang, Zhang, Liu, Awadallah, White, Burger, and Wang]{wu2023autogen}
Qingyun Wu, Gagan Bansal, Jieyu Zhang, Yiran Wu, Beibin Li, Erkang Zhu, Li~Jiang, Xiaoyun Zhang, Shaokun Zhang, Jiale Liu, Ahmed~Hassan Awadallah, Ryen~W. White, Doug Burger, and Chi Wang.
\newblock Autogen: Enabling next-gen llm applications via multi-agent conversation framework, 2023.
\newblock URL \url{https://arxiv.org/abs/2308.08155}.

\bibitem[Yao et~al.(2022)Yao, Zhao, Yu, Du, Shafran, Narasimhan, and Cao]{yao2022react}
Shunyu Yao, Jeffrey Zhao, Dian Yu, Nan Du, Izhak Shafran, Karthik Narasimhan, and Yuan Cao.
\newblock React: Synergizing reasoning and acting in language models, 2022.
\newblock URL \url{https://arxiv.org/abs/2210.03629}.

\bibitem[Zavaliangos-Petropulu et~al.(2019)Zavaliangos-Petropulu, Nir, Thomopoulos, Reid, Bernstein, Borowski, Jack, Weiner, and Thompson]{zavaliangos2019adni3}
Aikaterini Zavaliangos-Petropulu, Talia~M. Nir, Sophia~I. Thomopoulos, Richard~I. Reid, Matthew~A. Bernstein, Brian~J. Borowski, Clifford~R. Jack, Michael~W. Weiner, and Paul~M. Thompson.
\newblock Diffusion mri indices and their relation to cognitive impairment in brain aging: The updated multi-protocol approach in adni3.
\newblock \emph{Frontiers in Neuroinformatics}, 13:\penalty0 2, 2019.
\newblock \doi{10.3389/fninf.2019.00002}.

\end{thebibliography}

\end{document}